\documentclass[journal]{IEEEtran}
\ifCLASSINFOpdf
\else
\fi
\usepackage{graphicx} 
\usepackage{amsmath}
\usepackage{amssymb}
\usepackage{url}
\usepackage{xcolor}
\usepackage{soul}

\usepackage{comment}
\usepackage{amsmath,amssymb} 
\usepackage{color}
\usepackage{cite}
\usepackage{threeparttable}
\usepackage{comment}
\usepackage{adjustbox}
\usepackage{array, multirow}
\usepackage{colortbl}
\usepackage[percent]{overpic}

\usepackage{booktabs}
\usepackage{arydshln}

\def\eg{\textit{e.g.}}

\newcommand{\GG}{\mathbf{G}}
\newcommand{\ve}[1]{\mathbf{#1}} %

\begin{document}
%
\title{Towards Efficient Scene Understanding via Squeeze Reasoning}
%
\author{
Xiangtai Li,
Xia Li,
Ansheng You,
Li Zhang,
Guangliang Cheng,
Kuiyuan Yang,
Yunhai Tong,
Zhouchen Lin
\thanks{
	Xiangtai Li, Yunhai Tong are with the Key Laboratory of Machine Perception, School of EECS, Peking University.
	Li Zhang is with the School of Data Science, Fudan University.
	Kuiyuan Yang is with DeepMotion.
	Guangliang Cheng is with SenseTime.
	E-mail: \{lxtpku, yhtong\}@pku.edu.cn,
	lizhangfd@fudan.edu.cn,
	kuiyuanyang@deepmotion.ai,
    guangliangcheng2014@gmail.com
	}
}

%
%

\markboth{Journal of \LaTeX\ Class Files,~Vol.~14, No.~8, August~2015}%
{Shell \MakeLowercase{\textit{et al.}}: Bare Demo of IEEEtran.cls for IEEE Journals}
%



\maketitle

\begin{abstract}
	Graph-based convolutional model such as non-local block has shown to be effective for strengthening the context modeling ability in convolutional neural networks (CNNs). However, its pixel-wise computational overhead is prohibitive which renders it unsuitable for high resolution imagery. In this paper, we explore the efficiency of context graph reasoning and propose a novel framework called Squeeze Reasoning. Instead of propagating information on the spatial map, we first learn to squeeze the input feature into a channel-wise global vector and perform reasoning within the single vector where the computation cost can be significantly reduced. Specifically, we build the node graph in the vector where each node represents an abstract semantic concept. The refined feature within the same semantic category results to be consistent, which is thus beneficial for downstream tasks. We show that our approach can be modularized as an end-to-end trained block and can be easily plugged into existing networks. {Despite its simplicity and being lightweight, the proposed strategy allows us to establish the considerable results on different semantic segmentation datasets and shows significant improvements with respect to strong baselines on various other scene understanding tasks including object detection, instance segmentation and panoptic segmentation.} Code is available at \url{https://github.com/lxtGH/SFSegNets}.
	
\end{abstract}


%
\IEEEpeerreviewmaketitle

\section{Introduction}

\IEEEPARstart{C}{onvolutional} neural networks have proven to be effective and useful to learn visual representations in an end-to-end fashion with a certain objective task such as, semantic segmentation~\cite{fcn}, image classification~\cite{resnet}, object detection~\cite{faster-rcnn}, instance segmentation~\cite{maskrcnn} and panoptic segmentation~\cite{panoptic_seg}.
However, the effective receptive field~\cite{luo2016understanding} of CNNs grows slowly if we simply stack local convolutional layers. Thus, the limited receptive field prevents the model from taking all the contextual information into account and thus renders the model insufficiently covering all the regions of interest.

\begin{figure}[!t]
	
	\centering
	
	(a) Counting Pixel Results 
	{
		\begin{minipage}[t]{0.45\textwidth}
			\centering          
			\includegraphics[width=1.0\textwidth]{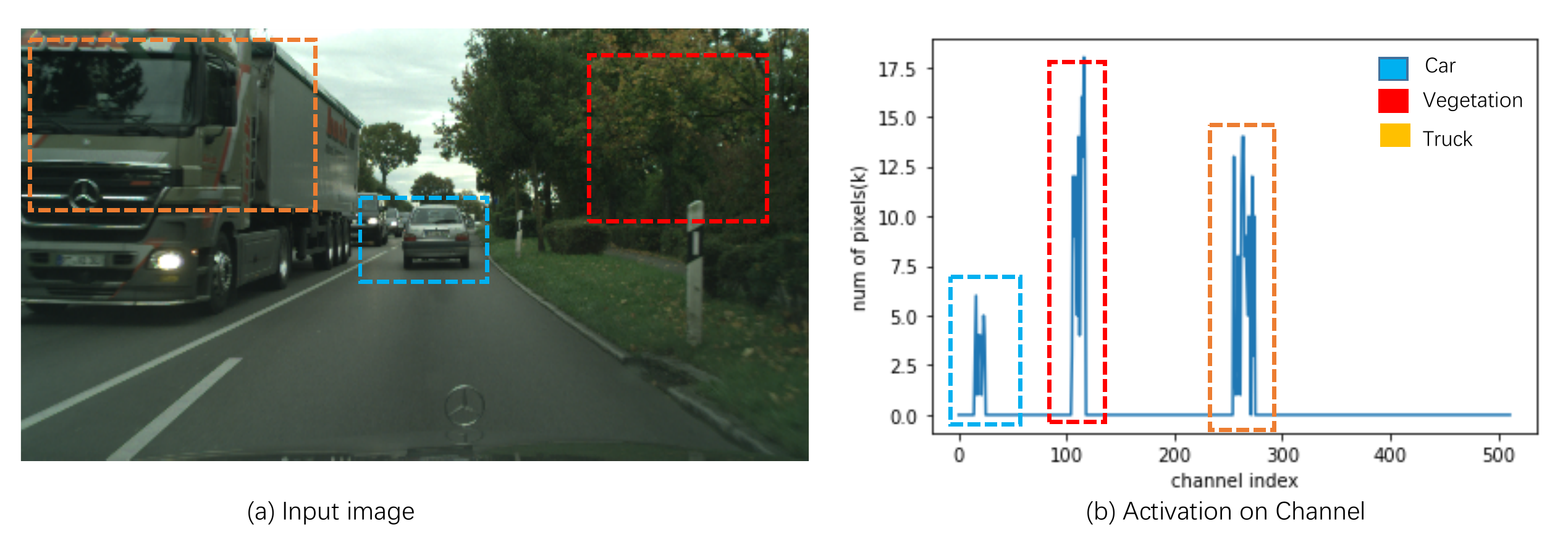} 
		\end{minipage}%
	}
	
	(b) Pipeline Ilustration
	{
		\begin{minipage}[t]{0.50\textwidth}
			\centering    
			\includegraphics[width=1.0\textwidth]{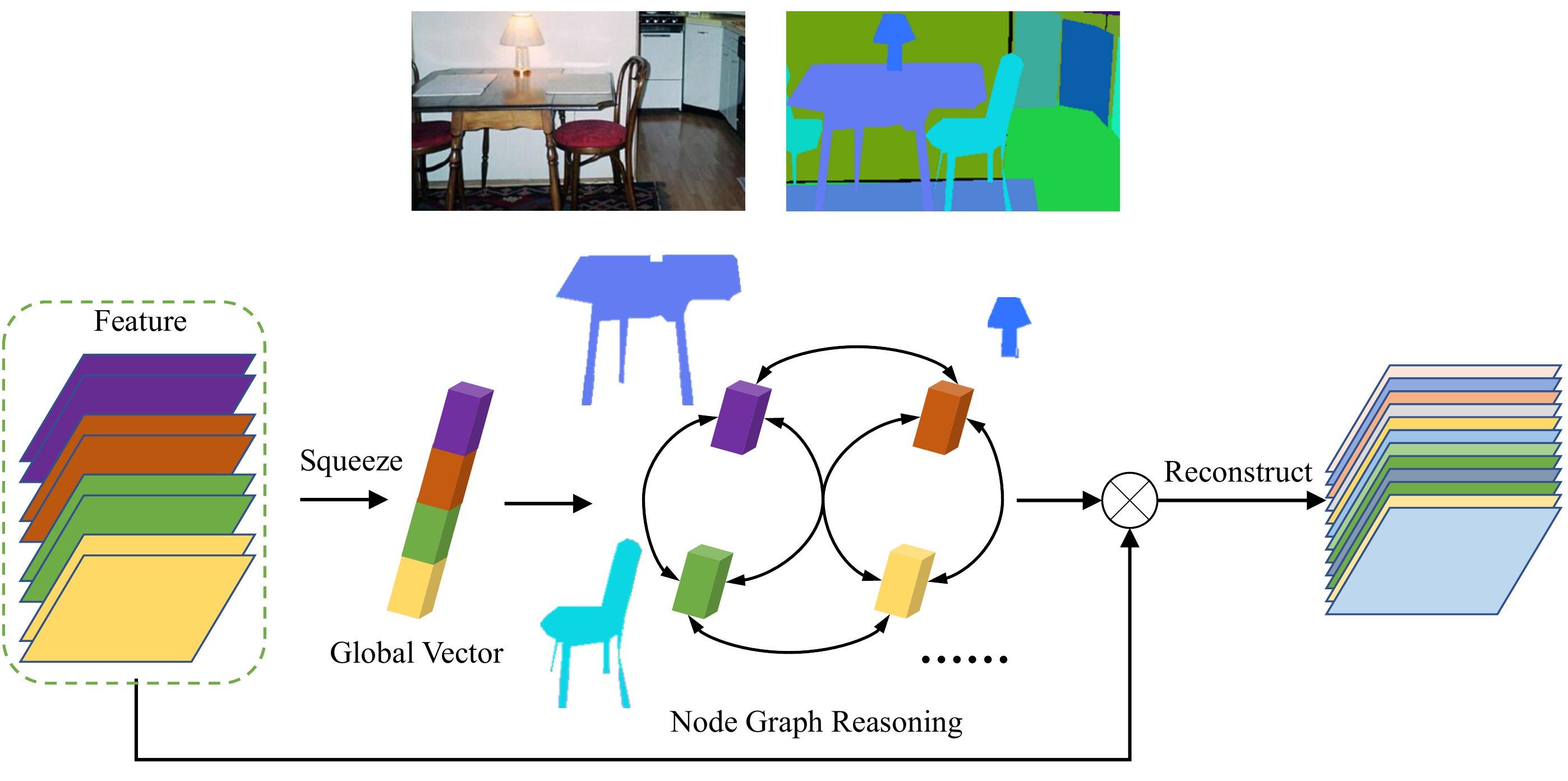}   
		\end{minipage}
	}%
	
	\caption{\footnotesize (a) Toy Experiment results by counting pixels given specific classes using trained model. (b) Illustration of our proposed module for semantic segmentation task. Best view in color and zoom in.} 
	\label{fig:teaser} 
\end{figure}

A broad range of prior research has investigated network architecture designs to increase the receptive field-of-view such as self-attention/Non-local~\cite{Nonlocal}, channel attention~\cite{senet}, and graph convolution network (GCN)~\cite{glore_gcn}. Although they have been shown to be effective in strengthening the representations produced by CNNs, 
modeling the inter-dependencies in a high-dimensional space prevents them from fully exploiting the sparse property required by the final classifier. Furthermore, they suffer from the prohibitively expensive computation overhead during training and inference,~\eg,  the large affinity matrix at each spatial position~\cite{senet} or channel position~\cite{glore_gcn}, which renders these methods unsuitable for high-resolution inputs. Although recent methods reduce such cost by involving fewer pixels ~\cite{ccnet} or selecting representative nodes~\cite{annnet}, their computation is still huge given high resolution image inputs.

\begin{table*}[t]
	\caption{Detailed results on Cityscapes Validation Set. In particular, our method can obtain a large improvement on large objects in the scene including train(24.1\%), truck(18.1 \%) and bus(17.4 \%).
	}
	\setlength{\tabcolsep}{2.4 pt}
	\begin{tabular}{ l | c c c c c c c c c c c c c c c c c c c | c c}
		\hline
		Method & road & swalk & build & wall & fence & pole & tlight & sign & veg. & terrain & sky & person & rider & car & truck & bus & train & mbike & bike & mIoU & GFlops\\
		\hline
		dilated FCN& 97.9 & 83.8 & 92.3 & 49.2 & 58.7 & 65.3 & 72.4 & 79.7 & 92.1 & 61.3 & 94.5 & 82.5 & 62.5 & 93.9 & \bf{66.0} & \bf 71.1 & \bf 51.0 & 66.0 & 77.8 & 74.8 & 241.05G \\
		+SR Head & 98.3 & 85.9 & 93.2 & 62.2 & 62.4 & 66.4 & 73.2 & 81.1 & 92.8 & 66.3 & 94.8 & 83.2 & 65.0 & 95.7 &\bf 84.1 & \bf89.6 &\bf 75.1 & 67.7 & 78.8 & 79.9(+5.1) & +3.64G \\
		\hline		
	\end{tabular}
	\label{tab:detailed_r50_res}
\end{table*}

Could we find another way to eliminate the limitation of high-cost spatial information while capturing global context information? 
We first carry out toy experiments using a pretrained Deeplabv3+ model~\cite{deeplabv3p}. We count the pixels on the final normalized feature (512 dimensions before classification) given ground truth masks whose activation values are beyond 0.8. As shown in Fig~\ref{fig:teaser}(a), we find different classes lie in different groups along channels sparsely. We only show three classes for simplicity.  This motivates us to build an information propagation module on channel solely where each group represents one specific semantic class while the cost of spatial resolution can be avoided. Inspired by SE-networks~\cite{senet}, we first squeeze the feature into a compact global vector and then perform reasoning operation on such compact vector.  Benefit from squeezing, the computation cost can be significantly reduced compared with previous works. The schematic illustration of our proposed method is shown in Fig~\ref{fig:teaser}(b). Compared with previous work modeling pair-wised affinity maps over the input pixels~\cite{ocnet,OCRNet,ccnet,encodingnet,EMAnet,RepGraph-2020}, our method is totally different by building node graph conditionally on the whole image statistics and also results in efficient inference. After reasoning, the most representative channels of input features can be seleceted and enhanced which solves the inconsistent segmentation results on large objects.

Our framework mainly contains three steps. First, we perform the node squeezing operation to obtain the global vector. This can be done by simply a global average pooling or using Hadamard product pooling to capture the second-order statistics. Then we carry out node reasoning by dividing such vector into different groups, and the inter-dependencies can be diffused through the reasoning process. Finally, we reconstruct the original feature map by multiplying the reasoned vector with the original input. Our approach can serve as a lightweight module and can be easily plugged into existing networks. Compared to Non-local~\cite{Nonlocal} or graph convolution network~\cite{glore_gcn}, which model the global relationship on feature spatial or channel dimension, our approach instead models the inter-dependencies on the squeezed global vector space, and notably, each node consists of a group of atom/channel. Therefore, our method uses substantially fewer floating-point operations and fewer parameters and memory costs. Moreover, our method achieves the best speed and accuracy trade-off on the Cityscapes test set, which shows its practical usage. In particular, our method achieves 77.5\% mIoU on Cityscapes test set while running at 65 FPS on single 1080-TI device.

\begin{figure}[t]
	\centering
	\includegraphics[width=0.95\linewidth]{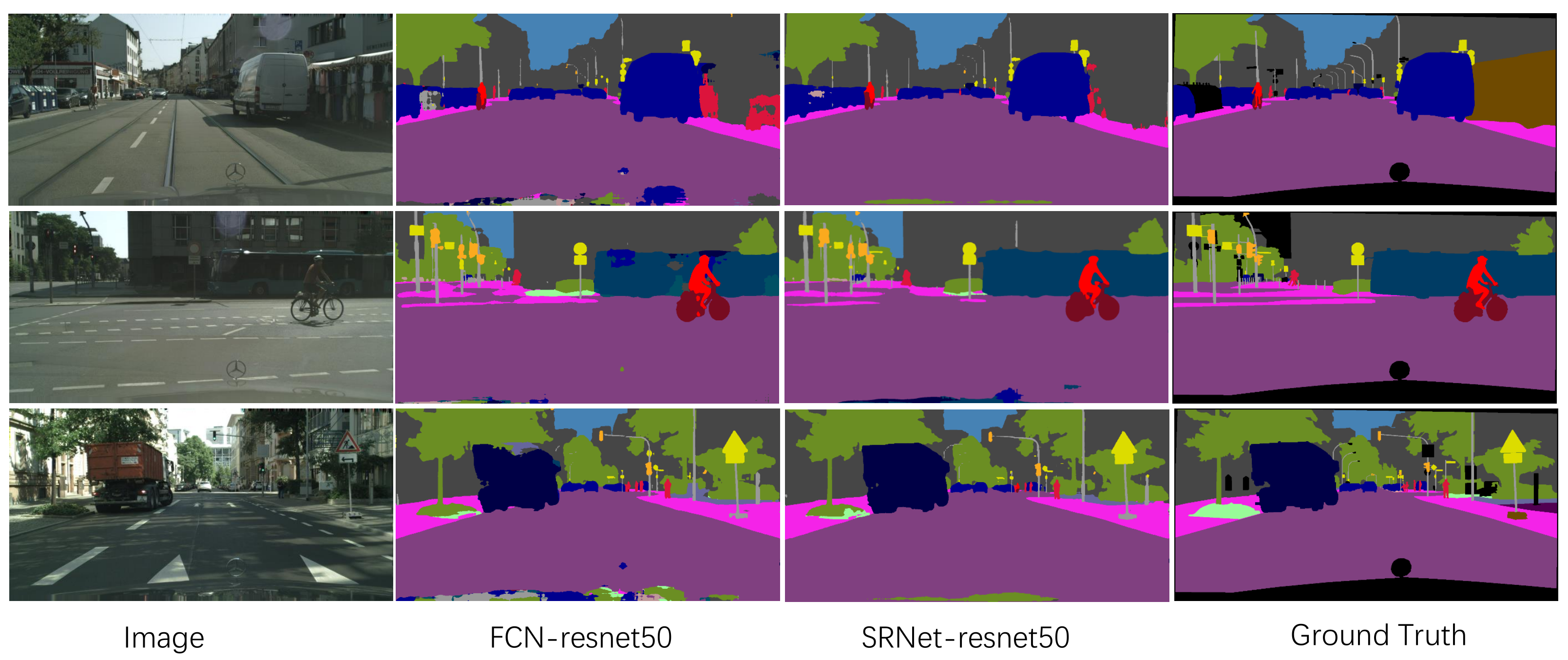}
	\caption{
		Visualization results on Cityscapes validation set. Best viewed in color and zoom in.}
	\label{fig:city_more_results}
\end{figure}

Besides its efficiency, our method is also verified to be effective on long-range context modeling. As shown in Tab.~\ref{tab:detailed_r50_res}, our method results in a significant gain on dilated FCN with ResNet50 backbone. In particular, the segmentation results of larger objects in the scene can be improved significantly by over 10\% mIoU per category. Moreover, our methods only require a few extra computation cost(1.5\% relative increase over the baseline models). Fig.\ref{fig:city_more_results} gives the visualization results on the corresponding models in Tab.~\ref{tab:detailed_r50_res}. As shown in that figure, the inconsistent noise on the train and truck can be removed by our proposed SR module.

Moreover, our proposed method is also verified to other tasks including instance level taks on instance segmentation and panoptic segmentation on COCO datasets~\cite{COCO_dataset}, image classification on ImageNet~\cite{imagenet}. Our method can obtain consistent improvements over mask-rcnn baseline models~\cite{maskrcnn} with negligible cost and considerable results on ImageNet classification with SE-net~\cite{senet}. More detailed information can be found in the experiment parts. Those experiments further demonstrate the generality of our proposed approach.

The contributions of this work are as follows:

(i) We propose a novel squeeze reasoning framework for highly efficient deep feature representation learning for scene understanding tasks.

(ii) An efficient node graph reasoning is introduced to model the inter-dependencies between abstract semantic nodes. This enables our method to serve as a lightweight and effective module and can be easily deployed in existing networks.

(iii)  Extensive experiments demonstrate that the proposed approach can establish new state-of-the-arts on four major semantic segmentation benchmark datasets including Cityscapes~\cite{Cityscapes}, Pascal Context~\cite{pcontext-data}, ADE20K~\cite{ADE20K} and Camvid~\cite{CamVid} while keeping efficiency, 
and show consistent improvement with respect to strong baselines on several scene understanding tasks with negligible cost. More experiments on different tasks throught datasets~\cite{COCO_dataset,imagenet} including instance segmentation and image classification prove the generality of proposed SR module.
\section{Related Work}
\label{sec:related}
In this section, we will review the related work in two aspects: global context aggregation and semantic segmentation.

\subsection{Global context aggregation}
Beyond the standard convolutional operator used for short-range modeling, many long-range operators are proposed to aggregate information from large image regions, even the whole image. Global Average Pooling (GAP)~\cite{resnet}, which bridges local feature maps and global classifiers, is widely used for long-range modeling. In Squeeze-and-Excitation network~\cite{senet}, GAP is used in more intermediate layers for coupling global information and local information more thoroughly. In Pyramid Pooling Module(PPM)~\cite{pspnet}, a pyramid of average pooling operators is used to harvest features. In addition to first-order statistics captured by GAP, bilinear pooling~\cite{BilinearCNN} extracts image-level second-order statistics as complementary of convolutional features. Besides pooling-based operators, generalized convolutional operators~\cite{atttention_aug} are also used for long-range modeling. Astrous convolution enlarges kernels by inserting zeros in between~\cite{dilation}, which is further used by stacking kernels with multiple astrous rates pyramidally~\cite{deeplabv1,deeplabv2,deeplabv3} or densely~\cite{denseaspp}. Deformable convolution~\cite{deformable,deformablev2} generalizes atrous convolution by learning the offsets for convolution sampling locations. Global average pooled features are concatenated into existing feature maps in ~\cite{parsenet}. In PSPNet~\cite{pspnet}, average pooled features of multiple window sizes including global average pooling are upsampled to the same size and concatenated together to enrich global information. The DeepLab series of papers~\cite{deeplabv1, deeplabv2, deeplabv3} propose atrous or dilated convolutions and atrous spatial pyramid pooling (ASPP) to increase the effective receptive field. DenseASPP~\cite{denseaspp} improves on \cite{deeplabv2} by densely connecting convolutional layers with different dilation rates to further increase the receptive field of network. In addition to concatenating global information into feature maps, multiplying global information into feature maps also shows better performance~\cite{encodingnet, cbam, cgnl, dfn}. In particular, EncNet~\cite{encodingnet} and SqueezeSeg~\cite{SqueezeSeg} use attention along the channel dimension of the convolutional feature map to account for global context such as the co-occurrences of different classes in the scene. CBAM~\cite{cbam} explores channel and spatial attention in a cascade way to learn task specific representation. 

Recently, advanced global information modeling approaches initiated from non-local network~\cite{Nonlocal} are showing promising results on scene understanding tasks. In contrast to convolutional operator where information is aggregated locally defined by local filters, non-local operators aggregate information from the whole image based on an affinity matrix calculated among all positions around the image. Using non-local operator, impressive results are achieved in OCNet~\cite{ocnet},CoCurNet~\cite{CoCurrentNet}, DANet~\cite{DAnet}, A2Net~\cite{a2net}, CCNet~\cite{ccnet} and Compact Generalized Non-Local Net~\cite{cgnl}. OCNet~\cite{ocnet} uses non-local bolocks to learn pixel-wise relationship while CoCurNet~\cite{CoCurrentNet} adds extra global average pooling path to learn whole scene statistic. DANet~\cite{DAnet} explores orthogonal relationships in both channel and spatial dimension using non-local operator. CCNet~\cite{ccnet} models the long range dependencies by considering its surrounding pixels on the criss-cross path through a recurrent way to save both computation and memory cost. Compact Generalized non-local Net~\cite{cgnl} considers channel information into affinity matrix. Another similar work to model the pixel-wised relationship is PSANet~\cite{psanet}. It captures pixel-to-pixel relations using an attention module that takes the relative location of each pixel into account. EMANet~\cite{2019Expectation} proposes to adopt expectation-maximization algorithm~\cite{EMalgorithm} for the self-attention mechanism.

Another way to get global representation is using graph convolutional networks, and do reasoning in a non-euclidean space~\cite{beyond_grids,spg_net_gcn,DGM_net} where messages are passing between each node before projection back to each position. Glore~\cite{glore_gcn} projects the feature map into interaction space using learned projection matrix and does graph convolution on projected fully connected graph. BeyondGrids~\cite{beyond_grids} learns to cluster different graph nodes and does graph convolution in parallel. SPGNet~\cite{spg_net_gcn} performed spatial pyramid graph reasoning while DGMNet~\cite{DGM_net} proposed dynamic graph reasoning framework for more efficient learning. In our work, a global vector squeezed from the whole image is organized as a small graph for reasoning, where each node contains rich global information. Thus reasoning is carried on a high level, which is more efficient and robust to noises than previous methods.

\subsection{Semantic segmentation}
Recent years have seen a lot of work on semantic segmentation using deep neural network. FCN~\cite{fcn} removes global information aggregation layers such as global average pooling layer and fully-connected layers for semantic segmentation. Later, FCN-based methods dominate the area of image semantic segmentation. We review related methods in two different settings: non-real-time methods for better segmentation results and real-time models for fast inference. The work~\cite{dilation} removed the last two downsample layers to obtain dense prediction and utilized dilated convolutions to enlarge the receptive field. Meanwhile, both SAC~\cite{sac} and DCN~\cite{deformable} improved the standard convolutional operator to handle the deformation and various scales of objects, which also enlarge the receptive fields of CNN opeartor. Several works~\cite{unet,upernet,panoptic_fpn,refinenet} adopted encoder-decoder structures that fuses the information in low-level and high-level layers to make dense prediction results. In particular, following such architecture design, GFFnet~\cite{xiangtl_gff}, CCLNet~\cite{ding2018context} and G-SCNN~\cite{gated-scnn} use gates for feature fusing to avoid noise and feature redundancy. CRF-RNN~\cite{crf_as_rnn} used graph model such CRF, MRF for semantic segmentation. AAF~\cite{aaf} used adversarial learning to capture and match the semantic relations between neighboring pixels in the label space. DenseDecoder~\cite{densedecoder} built multiple long-range skip connections on cascaded architecture. DPC~\cite{DPC} and auto-deeplab~\cite{auto-deeplab} utilized architecture search techniques to build multi-scale architectures for semantic segmentation. Besides, there are also several works aiming for real time application. ICNet~\cite{ICnet}, BiSegNet~\cite{bisenet} and SFNet~\cite{SFnet} were designed for real-time semantic segmentation by fusing multi scale inputs or feature pyramids. DFANet~\cite{dfanet} utilizes a light-weight backbone to speed up its network and proposes a cross-level feature aggregation to boost accuracy, while SwiftNet~\cite{swiftnet} uses lateral connections as the cost-effective solution to restore the prediction resolution while maintaining the speed. There are also specially designed video semantic segmentation works for boosting accuracy~\cite{Netwarp,GRFP_video} and saving inference time~\cite{DFF,Lowlatency_net}. 
Our proposed module can work in both real-time setting to obtain the best speed and accuracy trade-off due to its efficiency or non-real-time setting to achieve the better consistent segmentation results.

\section{Method}

In this section, we first review related works~\cite{senet,gcnpaper,Nonlocal} as preliminary knowledge. Then detailed description and formulation of our SR module are introduced. Finally, we elaborate on how to apply it to several different computer vision tasks.

\subsection{Preliminaries}

\begin{figure*}[!t]
	\centering  
	\includegraphics[width=1.0\linewidth]{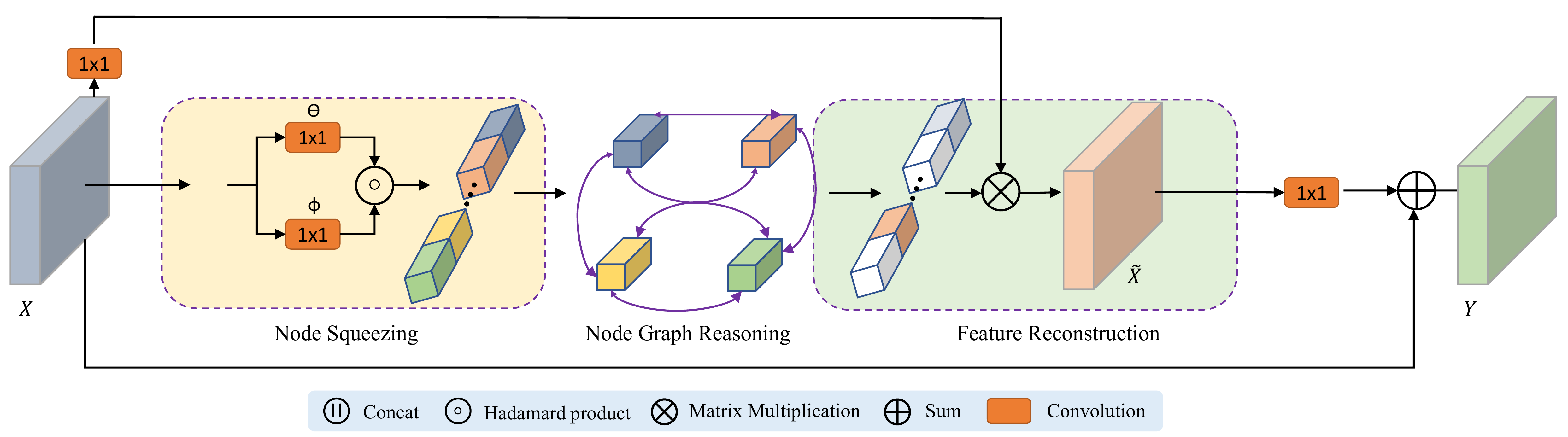}
	\caption{\footnotesize Schematic illustration of our proposed SR module. Our module contains three steps. Node Squeezing: squeeze the feature into separate nodes. Node Graph Reasoning: perform GCN reasoning in node space. Feature Reconstruction: reconstruct the feature by the reasoned global vector.
	}
	\label{fig:sr_overview}
\end{figure*}

\noindent
\textbf{Graph convolution}. 
Assume an input matrix $\mathbf{X} \in \mathbb{R}^{D \times N}$, where $D$ is the feature dimension and $N = H \times W$ is the number of locations defined on regular grid coordinates $\Omega=\{1,...,H\}\times\{1,...,W\}$.
In standard convolution, information is only exchanged among positions in a small neighborhood defined by the filter size~(typically $3\times 3$). 
In order to create a large receptive field and capture long-range dependencies, one needs to stack numerous layers after each other, as done in common architectures \cite{resnet}.
Graph convolution~\cite{gcnpaper}, is a highly efficient, effective and differentiable module that generalizes the neighborhood definition used in standard convolution and allows long-range information exchange in a single layer.
This is done by defining edges $\mathbb{E}$ among nodes $\mathbb{V}$ in a graph $\mathbb{G}$. 
Formally, the graph convolution is defined as
\begin{equation}\label{eq:gcn}    
\tilde{\mathbf{X}} =  \sigma(\mathbf{W} \mathbf{X} \mathbf{A}),
\end{equation}
where $\sigma(\cdot)$ is the non-linear activation function, $\mathbf{A}\in \mathbb{R}^{N\times N}$ is the adjacency matrix characterising the neighbourhood relations of the graph and  $\mathbf{W}\in\mathbb{R}^{D \times \tilde{D}}$ is the weight matrix. So the graph definition and structure play a key role in determining the information propagation.

\noindent
\textbf{Non-local network}. We describe non-local network~\cite{Nonlocal} in view of a fully connected graphical model. For a 2D input feature with the size of $C \times H \times W$, where $C$, $H$, and $W$ denote the channel dimension, height, and width respectively, it can be interpreted as a set of features, $\ve{X} = [\ve{x}_1,\ve{x}_2, \cdots, \ve{x}_N]^\intercal$, $\ve{x}_i\in\mathbb{R}^C$, where $N$ is the number of nodes (\eg $N=H \times W$), and $C$ is the node feature dimension. 
\begin{equation}\label{eq:nonlocal-matrix}
\begin{aligned}
\ve{\tilde{X}} = \delta(\ve{X_\theta}\ve{X_\phi}^\intercal)\ve{X_g} 
=\ve{A}(\ve{X})\ve{X_g},
\end{aligned}
\end{equation}
where $\ve{A}(\ve{X}) \in \mathbb{R}^{N \times N}$ indicates the affinity matrix, $X_\theta \in \mathbb{R}^{N \times C'}$, $X_\phi \in \mathbb{R}^{N \times C'}$, and $X_g \in \mathbb{R}^{N \times C'}$ which are projection matrix. In summary, according to Equ.~\ref{eq:gcn} and Equ.~\ref{eq:nonlocal-matrix} both the computation and affinity cost are highly dependent on the number of node $N$.  

\noindent
\textbf{`Squeeze' operation.}  
The `squeeze' operation is commonly used in networks for image classification, and adopted in SE-net~\cite{senet} to summarize the global contexts for intermediate layers for channel weights re-calibration. One simple implementation of this operation is the Global Average Pooling~(GAP).

\subsection{SR module formulation}
As discussed, pixels or nodes' choices are essential for reducing computation cost for both graph convolution models and non-local models. Recent works~\cite{ccnet, OCRNet} follow this idea to achieve less computation cost.  However, both the affinity memory and computation cost are still linearly dependent on the input resolution. In particular,  this will limit their usage for some applications such as road scene understanding with high-resolution image inputs. Different from their approaches, we propose a simple yet effective framework named Squeeze Reasoning. Our approach mainly contains three steps. First, we squeeze the input feature map into a compact global vector. We then split such vector into different groups or nodes and perform graph reasoning operation on such input node graphs. Finally, we reconstruct the original feature map by multiplying the reasoned vector back into the input feature. We specify the details of these three steps in the following parts. Fig.~\ref{fig:sr_overview} shows the detailed pipeline of our SR module.

\subsubsection{Node squeezing}

It is well known that the global vector describes whole image statistics, which is a key component in many modern convolutional network architectures for different tasks such as object detection, scene parsing and image generation. The simplest way to calculate the global vector is the global average pooling, which calculates the first-order whole image statistics.

Recent works~\cite{BilinearCNN,a2net} use the bilinear pooling to capture second-order statistics of features and generate global representations. Compared with the conventional average and max pooling, which only computes first-order statistics, bilinear pooling can better capture and preserve complex relations. In particular, bilinear pooling gives a sum pooling of second-order features from the outer product of all the feature vector pairs $(\mathbf{b}_{i}, \mathbf{c}_{i})$ within two input feature maps $\mathbf{B}$ and $\mathbf{C}$. It can be expressed as follows:
\begin{equation} \label{eq:bilinear}
\GG_{bilinear}\left(\mathbf{B}, \mathbf{C} \right) = \mathbf{B}\mathbf{C}^T = \sum_{i=1}^{HW} \mathbf{b}_{i} \mathbf{c}_{i}^T,
\end{equation}
where $\mathbf{B} = [\mathbf{b}_{1},\dots,\mathbf{b}_{i},\dots,\mathbf{b}_{HW}] \in \mathbb{R}^{C \times HW} $ and $\mathbf{C}=[\mathbf{c}_{1},\dots,\mathbf{c}_{i},\dots,\textbf{c}_{HW}] \in \mathbb{R}^{C \times HW}$. The output size is $C\times C$. 

To get a more compact vector for each node, in this paper, instead of generating outer product of all feature pairs from $\mathbf{b}_{i}$ and $\mathbf{c}_{i}$ within two input feature maps $\mathbf{B}$ and $\mathbf{C}$, we calculate the Hadamard product: 
\begin{equation} \label{eq:inner}
\GG_{global}\left( \mathbf{B}, \mathbf{C} \right) = \sum_{i=1}^{HW} \mathbf{b}_{i} \circ \mathbf{c}_{i},
\end{equation}
where $\circ$ means the Hadamard product operation. To be note that, we first reduce channel dimension of input feature $X$ by 1$\times$ 1 convolution layer into $\tilde{X}$ and then we perform pooling operation using Equ.~\ref{eq:inner} or simple global average pooling.

\subsubsection{Node graph reasoning}

To form a node graph, we divide vector $g$ into $k$ different groups with each group size of $d$ where $C = k \times d$.
We use the graph convolution to model the relationship between nodes and consider it a fully-connected graph. As for the transformation $\mathbf{W}$ in Equ.~\ref{eq:gcn}, we adopt a $1\times 1$ convolutional layer to implement it. Moreover, for the adjacency matrix $\mathbf{A}$, we will show by experiments that our Squeeze Reasoning mechanism is not sensitive to these choices, indicating that the generic graph reasoning behavior is the main reason for the observed improvements. We will describe two specific choices in the following part: 

\noindent
\textbf{Learned matrix}. We follow the same settings in GloRe~\cite{glore_gcn}, a simple choice of $\mathbf{A}$ is a $1 \times 1$ convolutional layer that can be updated by the general backpropagation. Similar to previous works~\cite{beyond_grids,glore_gcn}, we consider adopting the Laplacian matrix $\left( \mathbf{I} - \mathbf{A}_{g} \right)$ to propagate the node features over the graph, where the identity matrix $\mathbf{I}$ serves as a residual sum connection. In this setting, the Graph Reasoning can be formulated as follows:
\begin{equation}
\mathbf{G}_{output} = \sigma \left( \mathbf{W}_{g} \mathbf{G}_{input} \left(\mathbf{I} - \mathbf{A}_{g} \right)\right),
\label{eq:gcn1}
\end{equation}
where $\sigma$ is the ReLU operation.

\noindent
\textbf{Correlation matrix}. {Another choice is to adopt the self-attention mechanism for information exchange~\cite{Nonlocal,DAnet} where the correlation matrix (or dense affinity) is calculated by the projection of node feature itself,} by which the reasoning process can be written as follows:
\begin{equation}
\mathbf{G}_{output} = \sigma\left\{ \rho_g(\mathbf{G}_{input}) [\phi_g (\mathbf{G}_{input})^T \theta_g(\mathbf{G}_{input})]  \right\},
\label{eq:gcn2}
\end{equation}
where $\phi_g$, $\theta_g$ and $\rho_g$ are three $1 \times 1$ convolutions. $\phi_g$ and $\theta_g$ are named `query' and `key', respectively. The $\rho_g$ operation here, named `value', functions the same as the $\mathbf{W}_g$ in the `Learned matrix' mechanism. The results generated by operations inside the $[.]$ form the adjacency matrix $\mathbf{A}$. To be noted that, either reasoning process can be adopted in our framework and more detailed results can be referred to the experiment part.

\vspace{1mm}
\subsubsection{Feature reconstruction}

The final step is to generate the representation $\mathbf{R}$. To reconstruct the feature map, we first reshape the reasoned vector and multiply it with $X$ to highlight different channels according to the input scene where $\mathbf{\tilde{X}} = \mathbf{X} \mathbf{G}_{output}$. Then we adopt another 1$\times$1 convolution layers $W_{R}$ to project the $\tilde{X}$ into original shape.  Following the same idea of residual learning~\cite{resnet,Nonlocal}, we get the final output $\mathbf{Y}$ by adding original input $\mathbf{X}$. Then the feature map can be reconstructed as follows:
\begin{equation}
\mathbf{Y} =  \mathbf{W_{R} } \tilde{\mathbf{X}} + \mathbf{X}
\end{equation}
where $W_{R}$ is a learnable linear transformation and $Y$ is the final output feature.

\subsection{Analysis and discussion}

\noindent
\textbf{Relationship with Previous Operators:} Compared with the Non-local block~\cite{Nonlocal}, instead of affinity modeling on pixel-level, our SR extends the reasoning on channel dimension, which captures statistics of whole feature map space. Compared with the SE block~\cite{segnet}, our SR module captures more relational information and performs channel diffusion more efficiently than the fixed fully connected layers. Moreover, the experiment results show the advantages of our module.

\noindent
\textbf{Computation and memory analysis:} Compared with previous self-attention methods~\cite{Nonlocal,ccnet,a2net}, we compare our module computation in Tab.~\ref{tab:sr_computation} which shows our module is both lightweight on both computation and memory compared with previous methods. To be noted, we only consider the reasoning part in both computation cost and affinity memory. As shown in the last row, our module linearly depends on the input resolution in terms of computation and has no relation with the input resolution in terms of affinity memory. More analysis can be found in the experimental part.

\noindent
\textbf{Discussion with EncNet~\cite{encodingnet}:} {Despite both EncNet and our SRnet both adopt the channel attention framework~\cite{senet}, our method is different from EncNet. For squeezing process, EncNet learns an inherent dictionary to represent the semantic context of the dataset which is dataset dependent, while our method squeezes the entire feature map into one compact vector via learnable convolutions which is data dependent. The dataset dependent dictionary captures the class relation prior on specific dataset and thus it is hard for generalization. Our approach is more general and it can be a plug-in-play module. In addition, we have proven this point on many other tasks including object detection, instance segmentation and panoptic segmentation. For reasoning process, our method divides the global feature into different groups. Each group represents a latent class and the information diffusion is achieved by one graph convolution layer while EncNet adopts SE-like architecture. Our divided and reasoning process shows better results than SENet in Table-X. Moreover, EncNet also uses a semantic loss to achieve better results while our method only uses the cross-entropy loss as naïve FCN with much less tricks. Finally, we show the full advantages over different datasets under different settings over EncNet in the experimental part.}

\begin{table}[!t]
	\centering
	\caption{
			Time complexity comparison of non-local operations with our proposed SR module where $H$ and $W$ is spatial resolution and $C$ is channel dimension. $P$ is the order of Taylor expansion of kernel function in ~\cite{cgnl} and $KM = C/2$. Note that we ignore the channel reduction process since they are equal for computation cost.
	}
	\resizebox{0.5\textwidth}{!}{%
		\begin{tabular}{l|c|c}
			\hline
			{Methods} & {Computation} & {Affinity Memory} \\
			\hline
			Non-local~\cite{Nonlocal} & $O(C(HW)^2)$ &  $O((HW)^2)$\\
			A2Net~\cite{a2net} & $O(C^2(HW))$& $O(C^2)$\\
			CGNL~\cite{cgnl} & $O(CHWP))$& $O(P^2)$ \\
			CCNet~\cite{ccnet} & $O(CHW(H+W)$& $O(HW(H+W))$ \\
			DANet~\cite{DAnet} & $O(C(HW)^2 + HW(C)^2)$ &  $O((HW)^2+(C)^2)$ \\
			\hline 
			SRNet &$O(CHW + C)$& $O(K^2 + M^2)$ \\
			\hline
		\end{tabular}
	}
	\vspace{1mm}
	\label{tab:sr_computation}
\end{table}

\subsection{Network architecture}
The proposed SR module can be easily incorporated into the existing CNN architectures. We detail our network design in the task of semantic segmentation and instance level segmentation. 

\noindent
\textbf{Semantic Segmentation}: We adopt the Fully Convolution Networks~(FCNs)~\cite{fcn} as the backbone model. In particular, we choose ImageNet~\cite{imagenet} pretrained ResNet~\cite{resnet}, remove the last two down-sampling operations and adopt the multi-grid ~\cite{dilation} dilated convolutions. We remove the last two down-sampling operations and use the dilation convolutions instead to hold the feature maps from the last two stages 
$\frac{1}{8}$ of the input image. Concretely, all the feature maps in the last three stages have the same spatial size. Following the same  setting~\cite{ccnet,annnet}, we insert our proposed module between two 3 $\times$ 3 convolution layers (both layers output D = 512 channels ), which are appended at the end of the FCN. Following~\cite{pspnet}, our model has two supervisions: one after the final output of our model while another at the output layer of Stage4 as auxiliary cross-entropy loss. For real-time segmentation models, we choose DF-seg models~\cite{DF-seg-net} as a baseline and we replace their head with our SR module.

\noindent
\textbf{Instance level segmentation}: For instance segmentation and panoptic segmentation, We choose two-stage mask-rcnn-like architectures~\cite{maskrcnn,panoptic_fpn} We insert our module on the outputs of bottleneck in ResNet~\cite{resnet} layer4 for context modeling.

\section{Experiment} 
We verify the proposed module on four scene understanding tasks, including semantic segmentation, object detection, instance segmentation, panoptic segmentation and image classification. Our method outperforms several state-of-the-art methods on four benchmarks for semantic segmentation, including Cityscapes, ADE20K, Pascal Context and Camvid, with much less computation cost. Experiments on the other four vision tasks also demonstrate the effectiveness of the proposed module. All the experiments are under the same setting for each task and each dataset for a fair comparison.

\subsection{Ablation Experiments on Cityscapes dataset}

\noindent \textbf{Experimental settings on ablation studies:} We carry out detailed ablation studies and visual analysis on proposed approaches. We implement our method based on the PyTorch framework~\cite{pytorch}. For the Cityscapes dataset, following the same settings in PSPNet~\cite{pspnet}where momentum and weight decay coefficients are set to 0.9 and 5e-4 respectively, and ``poly'' learning rate schedule is used. For ablation studies, we choose ResNet-50 as the backbone where momentum and weight decay coefficients are set to 0.9 and 5e-4 respectively, and ``poly'' learning rate schedule is used which decays initial learning rate of 0.01 by multiplying $(1-\frac{\text{iter}}{\text{total}\_\text{iter}})^{0.9}$. Synchronized batch normalization is used~\cite{encodingnet}. For data augmentation, random cropping with size 769 and random left-right flipping are adopted. For the ablation studies, all models are trained by 50,000 iterations and evaluated by sliding-window crop inference. For the real time models, we use single scale full image inference.

\begin{table}[!t]
		\caption{\footnotesize Ablation study on the components of the proposed SR module. (a) Exploration on SR module design. \textbf{GHP}: Global Hadamard pooling. \textbf{FC}: Fully-connected layers. \textbf{GCN}: Reasoning with the graph convolution as Eq.~\ref{eq:gcn1}. \textbf{SA}: Reasoning using the self-attention mechanism as Eq.~\ref{eq:gcn2}. }
		\begin{minipage}{\dimexpr.90\linewidth}        
			\centering
			\resizebox{\textwidth}{!}{%
				\begin{tabular}{cc|ccc|cc}
					\hline
					\multicolumn{2}{c|}{Squeeze} & \multicolumn{3}{c|}{Reasoning} & \multirow{2}{*}{mIoU~(\%)} & \multirow{2}{*}{$\Delta$(\%)}\\ \cline{1-5}
					GAP           & GHP         & FC        & GCN       & SA             &   \\ \hline \hline
					-             & -           & -         & -         & -               & 74.8 & -  \\ \hline
					\checkmark    & -           & -         & -         & -              & 75.3 & +0.5  \\
					-             & \checkmark  & -         & -         & -            & 76.3 & +1.5  \\ \hline
					\checkmark    & -           & \checkmark & -         & -                & 76.6 & +1.8  \\
					-             & \checkmark  & \checkmark & -         & -                & 76.8 & +2.0  \\ \hline
					\checkmark    & -           & -         &\checkmark & -                 & \textbf{79.1} & +4.3 \\
					-             & \checkmark  & -         &\checkmark & -                & \textbf{\underline{79.9}} & \underline{+5.1}  \\
					-             & \checkmark  & -         & -         &\checkmark          & \textbf{79.5} & +4.7  \\
					\hline
				\end{tabular}
			}\par    
		\end{minipage}
	\label{tab:city_ablation}
\end{table}

\noindent \textbf{Ablation on SR framework design:} 
We first present a detailed analysis of each component of SR through ablation study and report results in Tab.~\ref{tab:city_ablation}(a). Comparing with the baseline, all SR versions equipped with different components achieve considerable improvements. By switching different squeezing operations, we find Global Hadamard Pooling (GHP) performs consistently better than Global Average Pooling (GAP) across differently used reasoning methods. Moreover, reasoning with Graph Convolutional Network and Self-Attention brings more improvements, even comparing with methods using global information from squeezing operation and further transformed by fully-connected layers(FC). Our segmentation models are trained under the best setting in this table. 

\begin{figure}[!t]
	\centering  
	\includegraphics[width=0.96\linewidth]{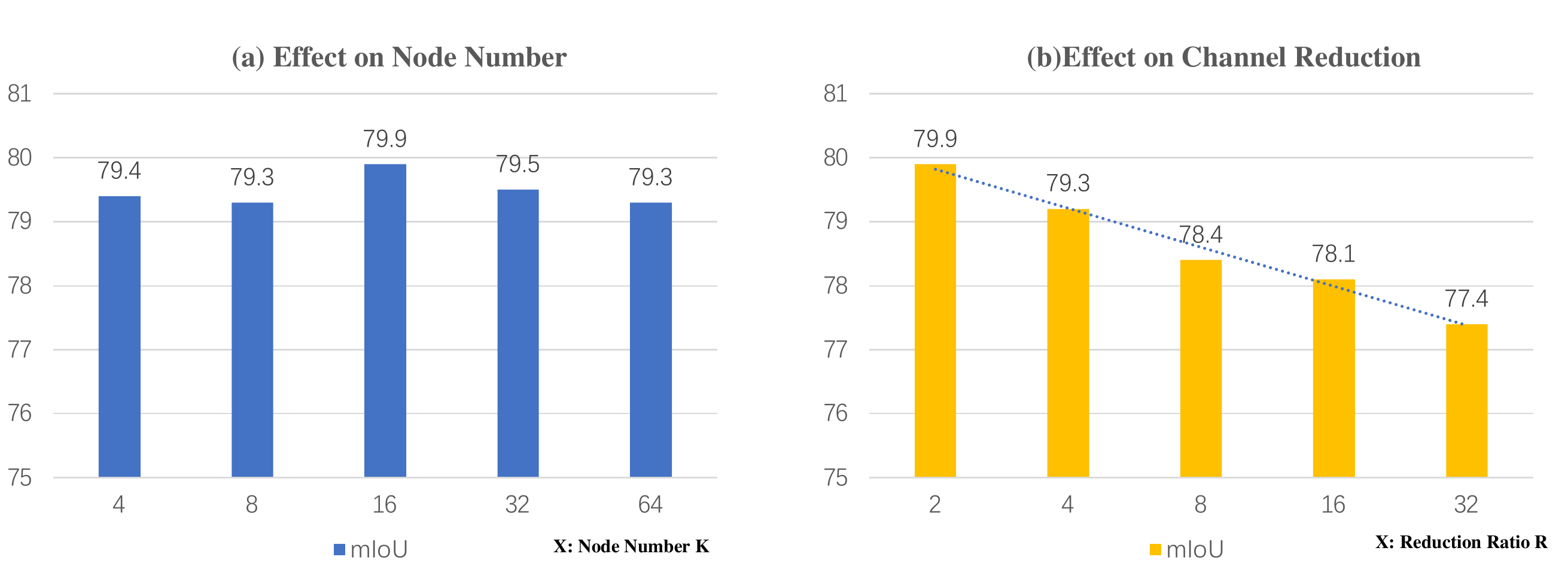}
	\caption{\footnotesize Ablation on hyper-parameter settings. (a) Effect on node number.
		(b) Effect on channel reduction.
	}
	\label{fig:ablation_on_node}
\end{figure}

\noindent \textbf{Ablation on hyper-parameter settings:}
To select the best hyper-parameter $K$, we also carry out an ablation study on the number of groups. To control independent variables, we fix $KM=C/2$, and only adjust K. We also explore the effect of channel reduction ratio with fixed $K$. The results are shown in Fig~\ref{fig:ablation_on_node}. From which, we can see that the selection of $K$ doesn't influence too much while reducing channel leads to inferior results. We set $K=16$ and ratio to 2 as default for the remaining experiments.

\begin{table}[!t]
	\centering
	\setlength\tabcolsep{1.1mm}
	\caption{Comparison experiments using different context modeling methods on the Cityscapes validation set, where dilated FCN~(ResNet-50) serves as the baseline method. The FLOPS and the Memory~(Mem) are computed over the input image of size $768 \times 768 \times 3$ and input feature map of size $96 \times 96 \times 2048$. For models other than ASPP and PSP, the overhead of the $3\times 3$ convolutions before and behind the modules are shown separately as the row $+2 * 3 \times 3$. \textbf{$\uparrow$} means the relative overheads over those of $+2 * 3 \times 3$. All the methods are evaluated under the same setting for the fair comparison.}
	\resizebox{0.45\textwidth}{!}{%
		\begin{tabular}{l|c|l|l|l}
			\hline
			Method & mIoU~(\%) & FLOPS & Params & Mem\\
			\hline  \hline
			dilated FCN & 74.8 & 241.05G & 23.63M & 3249M \\
			\hline
			+ASPP~\cite{deeplabv3} & 77.4 & +148.37G & +15.54M & +191.20M \\ 
			+PSP~\cite{pspnet} &  77.2 & +174.09G & +23.07M & +221.05M \\ 
			
			\hline \hline
			
			+$2*3\times 3$ & - & +108.74G & +11.80M & +108.00M \\ \hline
			+SE~\cite{senet} & 74.6 & \textbf{9.47M} $\uparrow$ & \textbf{0.03M$\uparrow$} & \textbf{18.88M$\uparrow$} \\
			+NL~\cite{Nonlocal} & 78.0 & 48.36G$\uparrow$ & 0.53M$\uparrow$ & 865.52M$\uparrow$ \\
			+A2Net~\cite{a2net} & 78.1 & 4.94G$\uparrow$ & 0.53M$\uparrow$ & 183.32M$\uparrow$ \\
			+CGNL~\cite{cgnl} & 78.2 & 4.91G$\uparrow$ & 0.53M$\uparrow$ & 201.94M$\uparrow$ \\
			+RCCA~\cite{ccnet} & \underline{78.5} & 11.55G$\uparrow$ & 0.53M$\uparrow$ & 394.28M$\uparrow$ \\
			{+Encoding~\cite{encodingnet} }& 77.5 & 12.31G$\uparrow$ & 0.59M$\uparrow$ & 257.12M$\uparrow$ \\
			{+ANN~\cite{annnet}} & 78.4 & 10.41G$\uparrow$ & 0.68M$\uparrow$ & 421.12M$\uparrow$ \\
			+EMAU~\cite{EMAnet} & 77.9 & 6.97G$\uparrow$ & 0.54M$\uparrow$ & 132.64M$\uparrow$ \\
			
			\hline
			\textbf{+SR~(GAP)} & \textbf{79.1} & \underline{2.43G$\uparrow$} & \underline{0.26M$\uparrow$} & \underline{82.31M$\uparrow$} \\
			\textbf{+SR~(GHP)} & \textbf{79.9} & 3.64G$\uparrow$ & 0.40M$\uparrow$ & 110.75M$\uparrow$ \\
			\hline  \hline
		\end{tabular}
	}
	\vspace{2mm}
	
	\label{tab:cityscapes_detail_compare}
	\vspace{0mm}
\end{table}

\begin{figure}[!t]
	\centering  
	\includegraphics[width=1.0\linewidth]{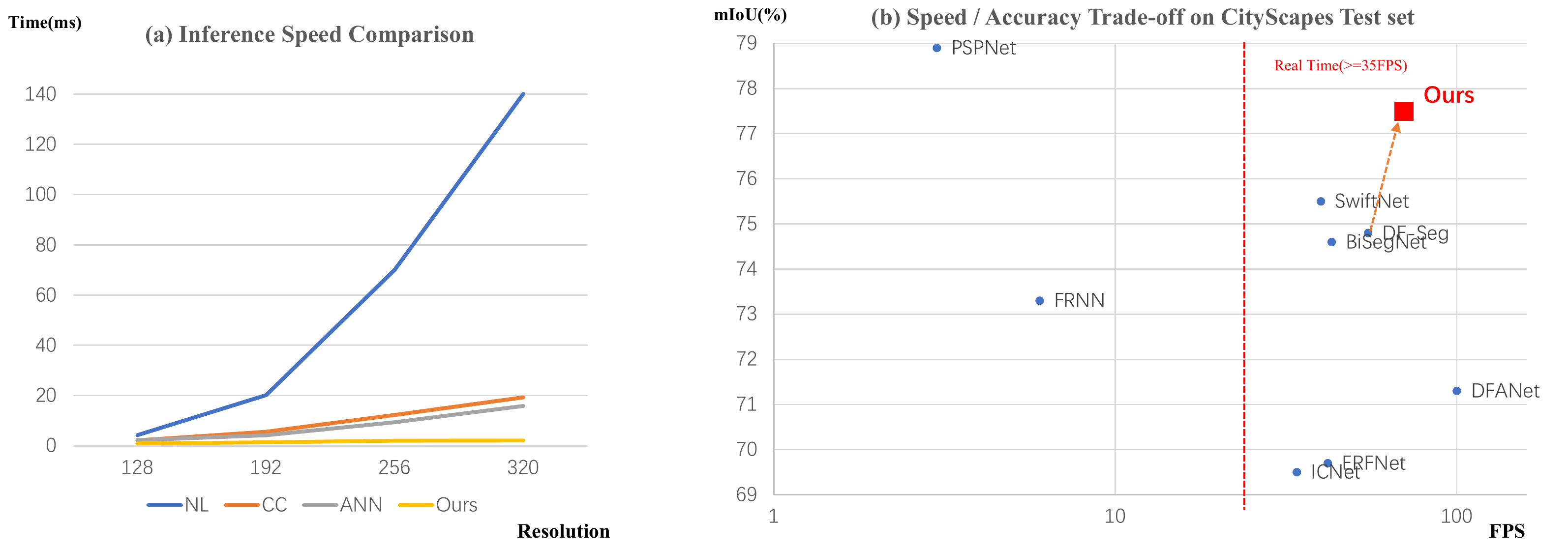}
	\caption{(a).Speed Comparison with Non-local and its variants. (b). Speed and Accuracy trade-off on Cityscapes Test Set for real time models. Best view it in color and zoom in.
	}
	\label{fig:speed_accuracy_comparison}
\end{figure}

\vspace{2mm}
\noindent \textbf{Comparisons with context aggregation approaches:}
\label{exp:analysis_efficiency}
In Tab.~\ref{tab:cityscapes_detail_compare}, we compare the performance of different context aggregation approaches, where SR achieves the best mIoU with ResNet-50 as the backbone. We give detailed and fair comparisons in terms of flops, parameters and memory cost. In particular, SR performs even better than all the non-local methods~\cite{ccnet,Nonlocal,cgnl,encodingnet}, which aggregates long-range contextual information in a pixel-wise manner. This indicates the effectiveness of cross-channel relationships in building compact and better representations with fewer computation FLOPS. Fig~\ref{fig:speed_accuracy_comparison}(a) gives inference time comparison with different resolution image inputs on V100-GPU, which shows the advantages with high-resolution image inputs.


\begin{figure*}
	\centering  
	\includegraphics[width=0.95\linewidth]{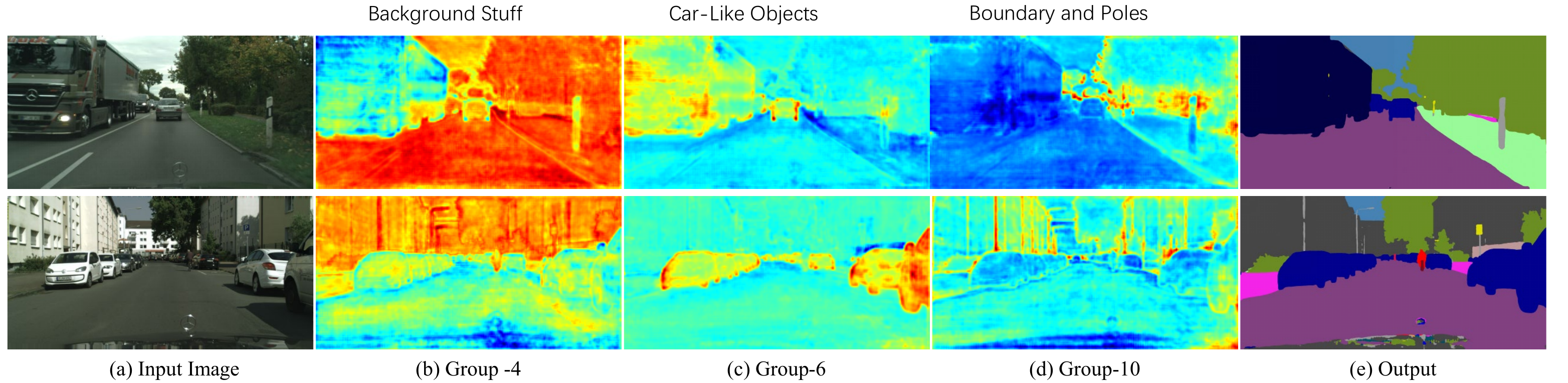}
	\caption{\footnotesize Visualization on learned group representation. We select the most salient channel from each node group. Such items capture specific concepts in the images. Best view in color and zoom in.
	}
	\label{fig:vis_learned_node}
\end{figure*}

\vspace{2mm}
\noindent \textbf{Visualization and analysis on learned node representation:} Here, we give visualization analysis on different channel activation on the reasoned feature map.  From Fig~\ref{fig:vis_learned_node}, we can see that each item corresponds to some abstract conceptions in the image. For example, the third-row item attends on the trucks and the cars while the second column shows that items focus on the stuff and background, and the fourth column items in group-10 are more sensitive to the small objects like poles and boundaries. 

\begin{figure}[!t]
	\centering  
	\includegraphics[width=0.95\linewidth]{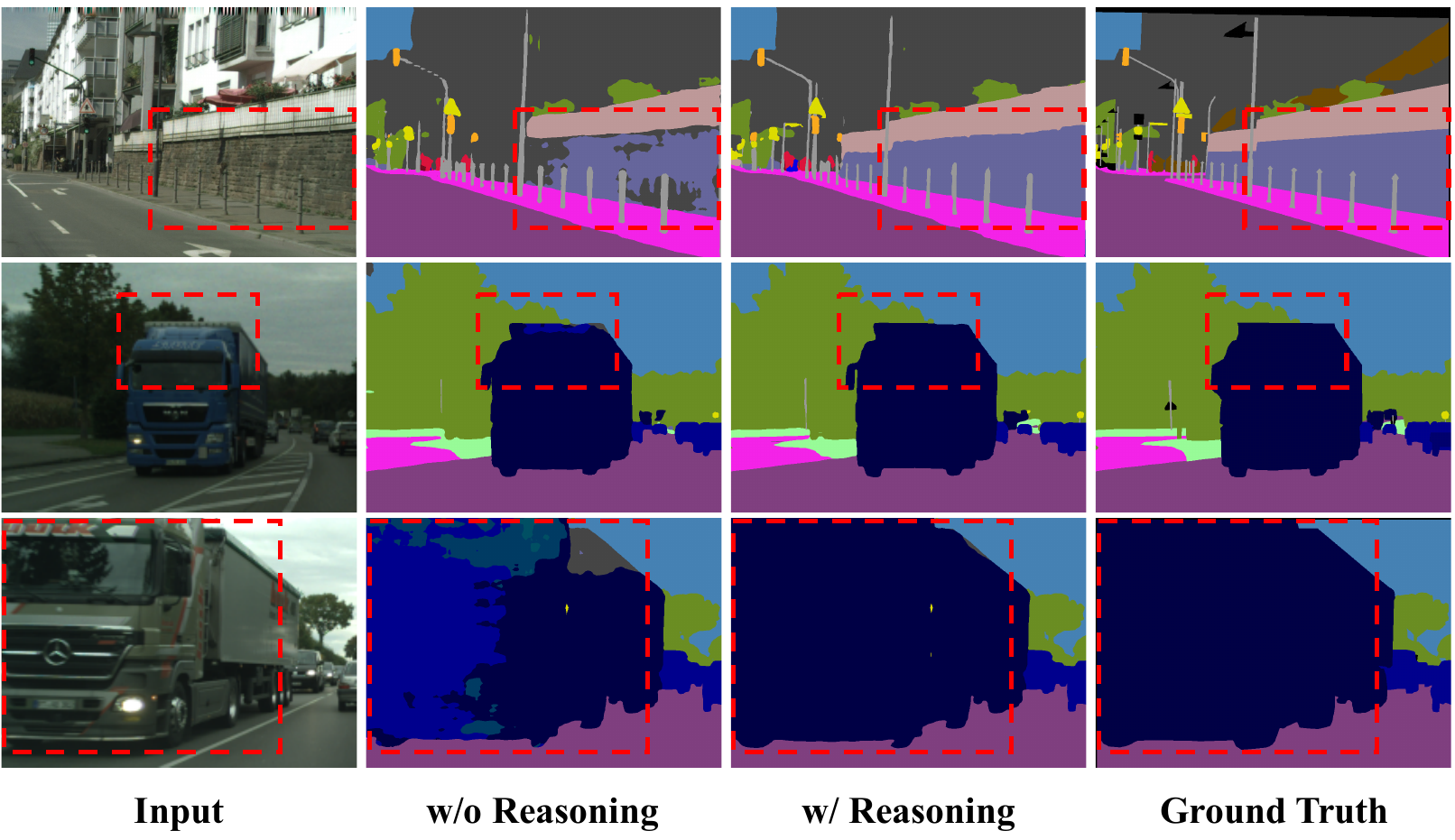}
	\caption{Comparison of our results from cropped images where dilated-FCN with GAP-
		Squeeze operation as the baseline model. Best view it in color and zoom in.}
	\label{fig:seg_results_comparison}
\end{figure}

\begin{figure}[!t]
	\centering  
	\includegraphics[width=0.95\linewidth]{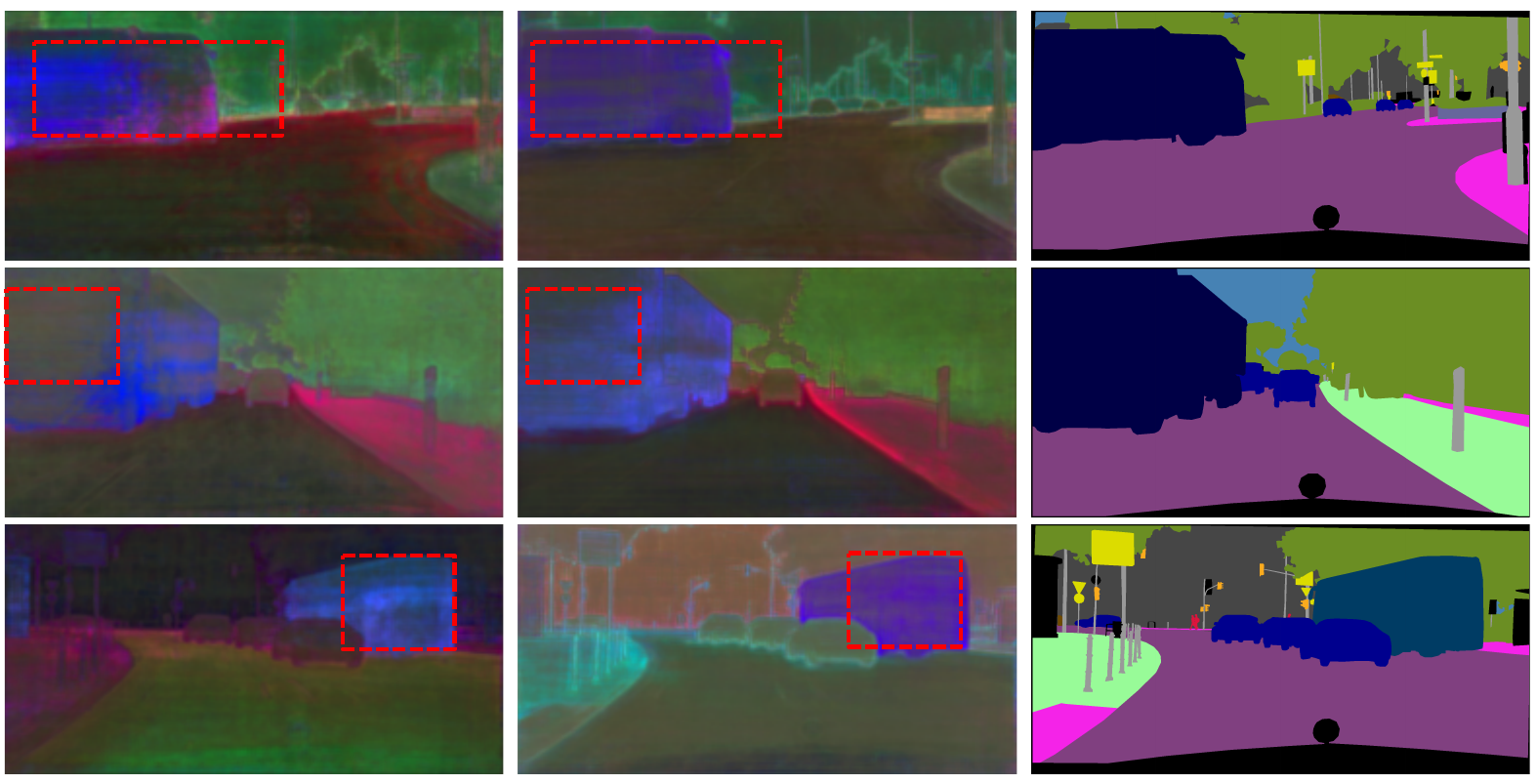}
	\caption{ The input and the output feature maps of the SR module. They are projected from 512-d to 3-d by PCA. Best view it in color and zoom in.}
	\label{fig:feature_visualization}
\end{figure}

\vspace{2mm}
\noindent \textbf{Visualization on predictions and feature maps:}
Fig.~\ref{fig:seg_results_comparison} compares segmentation results with and without reasoning. Without reasoning, pixels of large objects such as trucks and buses are often misclassified into similar categories due to ambiguities caused by limited receptive fields. The reasoning module resolves the above issue and delivers more consistent segmentation inside objects. Fig.~\ref{fig:feature_visualization} further investigates the effects of SR by directly comparing its input and output feature maps, where SR significantly improves the consistency of features inside objects, which is also the reason for consistent semantic map prediction. After SR, the features inner the objects have similar color and clearer boundaries shown the second column in Fig~\ref{fig:feature_visualization}.

\subsection{Experiments on Cityscapes in real-time settings}

\noindent{\textbf{Experiment settings:}}. Due to the efficiency of the proposed approach, we extend our method into real-time training settings. We mainly follow the DFANet~\cite{dfanet}. The networks with SR head are trained with the same setting, where stochastic gradient descent (SGD) with batch size of 16 is used as optimizer, with momentum of 0.9 and weight decay of 5e-4. All models are trained for 50K iterations with an initial learning rate of 0.01. As a common practice, the ``poly'' learning rate policy is adopted to decay the initial learning rate by multiplying $(1 -\frac{\text{iter}}{\text{total}\_\text{iter}})^{0.9}$ during training. Data augmentation contains random horizontal flip, random resizing with scale range of $[0.75, ~2.0]$, and random cropping with crop size of $1024 \times 1024$. During inference, we use the whole picture as input to report performance. To be more specific, we replace PPM head~\cite{pspnet} in DF-Seg-v2~\cite{DF-seg-net} with our module. Tab.~\ref{tab:cityscapes_sota_speed_acc} shows the results of our real-time model. Compared with baseline DFSeg~\cite{DF-seg-net}, Our method has similar parameters but with more accurate and faster speed.

\vspace{2mm}
\noindent \textbf{Comparisons with real time models on Cityscapes:} 
 We set a new record on \textbf{best speed and accuracy trade-off }in Cityscapes test set with 77.5 \%mIoU and 70 FPS shown in Fig.~\ref{fig:speed_accuracy_comparison} with full image resolution inputs ($1024 \times 2048$), which indicates the practical usage of our method. During inference, we use the whole picture as input to report performance. Tab~\ref{tab:cityscapes_sota_speed_acc} shows the results of our real-time model. Compared with previous real time models, our method achieves the best speed and accuracy trade-off. Compared with baseline DFSeg~\cite{DF-seg-net}, Our method has similar parameters but with more accurate and faster speed.

\begin{table}[!t]
	\centering
	\caption{Comparison on Cityscapes {\it test} set with state-of-the-art real-time models.  For fair comparison, input size is also considered, and all models use single scale inference.}
	\begin{threeparttable}
		\scalebox{0.90}{
			\begin{tabular}{l c c c c}
				\hline
				Method  &   InputSize & mIoU ($\%$) & \#FPS & \#Params \\
				\hline
				ESPNet~\cite{ESPNet} & $512 \times 1024$ & 60.3 & 132 & 0.4M \\
				ESPNetv2~\cite{ESPNetv2} &  $512 \times 1024$ & 62.1 & 80 & 0.8M \\
				ERFNet~\cite{ERFNet} & $512 \times 1024 $& 69.7 & 41.9 & - \\
				BiSeNet(ResNet-18)~\cite{bisenet} &  $768 \times 1536$ &  $74.6$ & 43 & 12.9M  \\
				BiSeNet(Xception-39)~\cite{bisenet} &  $768 \times 1536$ &  $68.4$ & 72 & 5.8M  \\ 
				ICNet~\cite{ICnet} & $1024 \times 2048$ &  69.5 & 34 & 26.5M \\
				DFv1\cite{DF-seg-net} & $1024 \times 2048$ & 73.0 & 80 & 9.37M  \\
				
				SwiftNet~\cite{swiftnet} & $1024 \times 2048$  & 75.5 & 39.9 & 11.80M \\
				SwiftNet-ens~\cite{swiftnet} & $1024 \times 2048$  & 76.5 & 18.4 & 24.7M \\
				DFANet~\cite{dfanet} & $1024 \times 1024$ & 71.3 & 100 & 7.8M \\
				CellNet~\cite{fast_cell_search_seg} & $768 \times 1536$ & 70.5  & 108 & -\\
				DFv2(baseline)~\cite{DF-seg-net} & $1024 \times 2048$ & \underline{74.8} & \underline{55} & \underline{18.83M}  \\
				\hline
				SRNet(DFv2) & $1024 \times 2048$ & $\bf{77.5}$ & \bf{65} & \textbf{18.87M} \\
				\hline
			\end{tabular}
		}
		
		\label{tab:cityscapes_sota_speed_acc}
	\end{threeparttable}
\end{table}

\vspace{2mm}
\subsection{Comparisons with state-of-the-art methods in non-real-time setting:}
In this section, we compare our method with state-of-the-art methods on four semantic segmentation benchmarks using multi scale inference setting. Without bells and whistles, our method outperforms several state-of-the-art models while costing less computation.

\vspace{1mm}
\noindent \textbf{Results on Cityscapes :}
We train our model for 120K iterations using only the finely annotated data (trainval set), online hard negative mining is used following~\cite{DAnet}. Multi-scale and horizontal flip testing is used as previous works~\cite{ccnet}. Tab.~\ref{tab:road_scene_seg_results}(a) compares the results, where our methods achieves \textbf{82.2\%} mIoU and outperforms all previous state-of-the-art models by a large margin. In particular, our method is 0.7\% mIoU higher than DANet~\cite{DAnet}, which uses non-local-like operator and is much efficient in both computation and memory cost due to the design of squeeze and reasoning. Our ResNet-50 based model achieves 81.0\% mIoU and outperforms DenseASPP~\cite{denseaspp} by 0.4\% with much larger backbone~\cite{densenet}, which shows the effectiveness of our method. After replacing stronger backbone Wider-ResNet~\cite{resnet38}, we achieve \textbf{83.3\%}mIoU with \textbf{only} fine annotated data, which outperforms previous state-of-the-art methods by a large margin. Note that we follow the G-SCNN~\cite{gated-scnn} setting by using Deeplabv3+ based Wider-ResNet pretrained on Mapillary~\cite{mapillary}. The detailed results are shown in Tab.~\ref{tab:cityscapes_results_detail_fine} for reference.

\begin{table*}
	\centering 
	\small
	\caption{
		Per-category results on the Cityscapes test set compared with accurate models. Note that all the models are trained with only fine annotated data. Our method with ResNet101 backbone outperforms existing approaches on \textbf{15} out of 19 categories, and achieves \textbf{82.2\%} mIoU. ss means single scale inference.}
	\addtolength{\tabcolsep}{0pt}
	\resizebox{\textwidth}{!}{
		\begin{tabular}{ l | c c c c c c c c c c c c c c c c c c c | c}
			\hline
			Method & road & swalk & build & wall & fence & pole & tlight & sign & veg. & terrain & sky & person & rider & car & truck & bus & train & mbike & bike & mIoU \\
			\hline
			DUC-HDC~\cite{hdc-duc} & 98.5 & 85.5 & 92.8 & 58.6 & 55.5 & 65.0 & 73.5 & 77.8 & 93.2 & 72.0 & 95.2 & 84.8 & 68.5 & 95.4 & 70.9 & 78.7 & 68.7 & 65.9 & 73.8 & 77.6 \\
			ResNet38~\cite{resnet38} & 98.5 & 85.7 & 93.0 & 55.5 & 59.1 & 67.1 & 74.8 & 78.7 & 93.7 & 72.6 & 95.5 & 86.6 & 69.2 & 95.7 & 64.5 & 78.8 & 74.1 & 69.0 & 76.7 & 78.4 \\
			PSPNet~\cite{pspnet} & 98.6 & 86.2 & 92.9 & 50.8 & 58.8 & 64.0 & 75.6 & 79.0 & 93.4 & 72.3 & 95.4 & 86.5 & 71.3 & 95.9 & 68.2 & 79.5 & 73.8 & 69.5 & 77.2 & 78.4\\
			AAF~\cite{aaf} & 98.5 & 85.6 & 93.0 & 53.8 & 58.9 & 65.9 & 75.0 & 78.4 & 93.7 &
			72.4 & 95.6 & 86.4 & 70.5 & 95.9 & 73.9 & 82.7 & 76.9 & 68.7 & 76.4 & 79.1 \\
			SegModel~\cite{segmodel} & 98.6 & 86.4 & 92.8 & 52.4 & 59.7 & 59.6 & 72.5 & 78.3 & 93.3 & 72.8 & 95.5 & 85.4 & 70.1 & 95.6 & 75.4 & 84.1 & 75.1 & 68.7 & 75.0 & 78.5 \\
			DFN~\cite{dfn} & - & - & - & - & - & - & - & - & - & - & - & - & - & - & - & - & - & - & - & 79.3 \\
			BiSeNet~\cite{bisenet} & - & - & - & - & - & - & - & - & - & - & - & - & - & - & - & - & - & - & - & 78.9 \\
			PSANet~\cite{psanet} & - & - & - & - & - & - & - & - & - & - & - & - & - & - & - & - & - & - & - & 80.1 \\
			DenseASPP~\cite{denseaspp}  & 98.7 & 87.1 & 93.4 & 60.7 & 62.7 & 65.6 & 74.6 & 78.5 & 93.6 & 72.5 & 95.4 & 86.2 & 71.9 & 96.0 & \textbf{78.0} & \textbf{90.3} & 80.7 & 69.7 & 76.8 & 80.6 \\
			BFPNet~\cite{BAFPNet} & 98.7 & 87.1 & 93.5 & 59.8 & \textbf{63.4} & 68.9 & 76.8 & 80.9 & 93.7 & 72.8 & 95.5 & 87.0 & 72.1 & 96.0 & 77.6 & 89.0 & 86.9 & 69.2 & 77.6 & 81.4 \\ 
			DANet~\cite{DAnet} & 98.6 & 87.1 & 93.5 & 56.1 & 63.3 & 69.7 & 77.3 & 81.3 & 93.9 & 72.9 & 95.7 & 87.3 & 72.9 & 96.2 & 76.8 & 89.4 & 86.5 & \textbf{72.2} & 78.2 & 81.5 \\
			\hline
			SRNet(ss)-ResNet101 & 98.7 & 86.7 & 93.5 & 60.9 & 61.7 & 68.3 & 76.6 & 79.9 & 93.7 & 72.4 & 95.8 & 86.9 & 72.3 & 96.1 & 76.1 & 87.2 & 88.2 & 70.3 & 77.5 & 81.2 \\
			SRNet-ResNet101 & \textbf{98.8} & \textbf{88.0} & \textbf{93.9} & \textbf{64.6} & 63.3 & \textbf{71.5} & \textbf{78.9} & \textbf{81.8} & \textbf{93.9} & \textbf{73.7} & \textbf{95.8} & \textbf{87.9} & \textbf{74.5} & \textbf{96.4} & 72.4 & 88.2 & \textbf{86.2} & 72.0 & \textbf{79.0} & \textbf{82.2} \\
			\hline
			SRNet-WiderResNet & 98.8 & 87.9 & 94.2 & 65.2 & 66.0 & 72.4 & 77.8 & 81.5 & 94.1 & 75.0 & 96.5 &87.9 & 74.9 & 96.4 & 78.4 & 93.4 & 88.1 & 73.9 & 78.6 & 83.3 \\
			\hline
			
		\end{tabular}
	}
	\vspace{2mm}
	
	\label{tab:cityscapes_results_detail_fine}
\end{table*}

\vspace{1mm}
\noindent
\textbf{Results on CamVid:} is another road driving dataset. 
Camvid involves 367 training images, 101 validation images and 233 testing images with resolution of $960 \times 720$.
We use a crop size of 640 and training with 100 epochs. 
The results are shown in Tab~\ref{tab:road_scene_seg_results}(b). We report results with ResNet-50 and ResNet-101 backbone. 
With ResNet-101 as backbone, our method achieves \textbf{78.3\%} mIoU, outperforming the state-of-the-art approach~\cite{VideoGCRF} by a large margin~(3.1\%).


\begin{table}[!t]
	\centering
	\caption{ Comparison with the state-of-the-art methods on road-driving scene datasets including Cityscapes and Camvid.}
	\begin{minipage}{0.5\textwidth}
		
		\begin{minipage}{\dimexpr.9\linewidth}        
			\centering
			\resizebox{0.9\textwidth}{!}{%
				\begin{tabular}{l|c|c}
					\hline
					Method & Backbone & mIoU~(\%)  \\
					\hline \hline
					DFN~\cite{dfn} & ResNet-101 &  79.3 \\  
					CFNet~\cite{CoCurrentNet} & ResNet-101 &  79.6 \\
					DenseASPP~\cite{denseaspp}  & DenseNet-161 & 80.6 \\
					GloreNet~\cite{glore_gcn} & ResNet-101 & 80.9 \\
					BAFPNet~\cite{BAFPNet} & ResNet-101 & 81.4 \\
					CCNet~\cite{ccnet} & ResNet-101 & 81.4 \\
					ANNet~\cite{annnet} & ResNet-101 & 81.3 \\
					DANet~\cite{DAnet}  & ResNet-101 & 81.5 \\ 
					RGNet~\cite{RepGraph-2020} & ResNet-101 & 81.5 \\
					DGMN~\cite{DGM_net} & ResNet-101 & 81.6 \\
					OCRNet~\cite{OCRNet} & ResNet-101 &  81.8 \\
					\hline
					\textbf{SRNet} & ResNet-50 & \textbf{81.0}  \\
					\textbf{SRNet} & ResNet-101 & \textbf{82.2}  \\
					\hline
					G-SCNN~\cite{gated-scnn} & Wider-ResNet-38 & 82.8 \\
					\textbf{SRNet} & Wider-ResNet-38 & \textbf{83.3}  \\
					\hline \hline
				\end{tabular}        
			}\par
			\vspace{2mm}
			{(a) Results on the Cityscapes test set. All methods use only finely annotated data.}    
		\end{minipage}
		\vspace{2mm}
			
		 \begin{minipage}{\dimexpr.9\linewidth}
			\centering
			\resizebox{0.9\textwidth}{!}{%
				\begin{tabular}{l|c|c}
					\hline
					Method & Backbone & mIoU~($\%$)  \\
					\hline \hline
					SegNet~\cite{segnet} & VGG-16 & 60.1 \\
					RTA~\cite{RTA} & VGG-16 & 62.5 \\
					BiSeg~\cite{bisenet} & ResNet-18 & 68.7 \\ 
					PSPNet~\cite{pspnet} & ResNet-50 & 69.1 \\ 
					\textbf{SRNet} & ResNet-50 & \textbf{74.3} \\
					\hline
					DilatedNet~\cite{dilation} & ResNet101 & 65.3 \\
					Dense-Decoder~\cite{densedecoder} & ResNext-101 & 70.9 \\
					BFP~\cite{BAFPNet} & ResNet101 & 74.1 \\
					VideoGCRF~\cite{VideoGCRF} &ResNet101 & 75.2\\
					\hline
					\textbf{SRNet} & ResNet-101 & \textbf{78.3} \\
					\hline \hline
				\end{tabular}        
			}\par
			\vspace{2mm}
			{(b) Results on the CamVid test set.}    
		\end{minipage}

\end{minipage}
	\label{tab:road_scene_seg_results}
\end{table}

\begin{table}[!t]
	\caption{ Comparison with the state-of-the-art methods on more scene parsing datasets including ADE20k and Pascal Context.}
	\begin{minipage}{0.5\textwidth}
		\begin{minipage}{\dimexpr.9\linewidth}        
			\centering
			\resizebox{0.8\textwidth}{!}{%
				\begin{tabular}{l|c|c}
					\hline
					Method & Backbone & mIoU (\%)   \\
					\hline \hline
					EncNet~\cite{context_encoding} & ResNet-50  & 49.2  \\
					DANet~\cite{DAnet}& ResNet-50  & 50.1   \\
					\textbf{SRNet} & ResNet-50 & \bf 50.8  \\
					\hline
					EncNet~\cite{context_encoding} & ResNet-101   &51.7  \\
					Ding~\textit{et al.}~\cite{ding2018context} & ResNet-101  &51.6  \\
					DANet~\cite{DAnet}& ResNet-101  & 52.6   \\
					SGR~\cite{SGR_gcn} & ResNet-101 & 52.5 \\
					ANN~\cite{annnet} & ResNet-101 & 52.8 \\
					BAFPNet~\cite{BAFPNet} & ResNet-101 & 53.6 \\
					EMANet~\cite{EMAnet} & ResNet-101 & 53.1 \\
					GFFNet~\cite{xiangtl_gff} & ResNet-101 & 54.2 \\
					CFNet~\cite{zhang2019co} & ResNet-101 & 54.1 \\
					SPNet~\cite{hou2020strip} & ResNet-101 & 54.5 \\
					APCNet~\cite{apc_net} & ResNet-101 &\bf 54.7\\
					\hline
					\textbf{SRNet} & ResNet-101 &\bf 54.7\\
					\hline \hline
				\end{tabular}
			}\par
			\vspace{2mm}
			{(a) Results on Pascal Context dataset.}    
		\end{minipage}
		
		\begin{minipage}{\dimexpr.9\linewidth}
			\centering
			\resizebox{0.8\textwidth}{!}{%
				\begin{tabular}{l|c|c}
					\hline
					Method & Backbone & mIoU~(\%)  \\
					\hline \hline
					PSPNet~\cite{pspnet} & ResNet-50 &  42.78 \\ 
					PSANet~\cite{psanet} & ResNet-50 &  42.97  \\
					UperNet~\cite{upernet} & ResNet-50 &  41.55 \\
					EncNet~\cite{context_encoding} & ResNet-50 &  41.11 \\
					GCUNet~\cite{beyond_grids} & ResNet-50 & 42.60  \\
					\textbf{SRNet} & ResNet-50 & \textbf{43.42} \\ \hline
					PSPNet~\cite{pspnet} & ResNet-101 &  43.29\\ 
					PSANet~\cite{psanet} & ResNet-101 &  43.77\\ 
					SAC~\cite{sac} & ResNet-101 & 44.30 \\
					EncNet~\cite{context_encoding} & ResNet-101 & 44.65    \\ 
					GCUNet~\cite{beyond_grids} & ResNet-101 & 44.81\\
					ANN~\cite{annnet} & ResNet-101 & 45.24 \\
					\textbf{SRNet} & ResNet-101 & \textbf{45.53} \\
					\hline \hline
				\end{tabular}        
			}\par
			\vspace{2mm}
			{(b) Results on the ADE20K dataset.}
		\end{minipage}
	\end{minipage}
	\vspace{-5mm}
	\label{tab:seg_results_other}
\end{table}

\begin{figure}[t]
	\centering
	\vspace{-0.5\baselineskip}
	\includegraphics[width=0.95\linewidth]{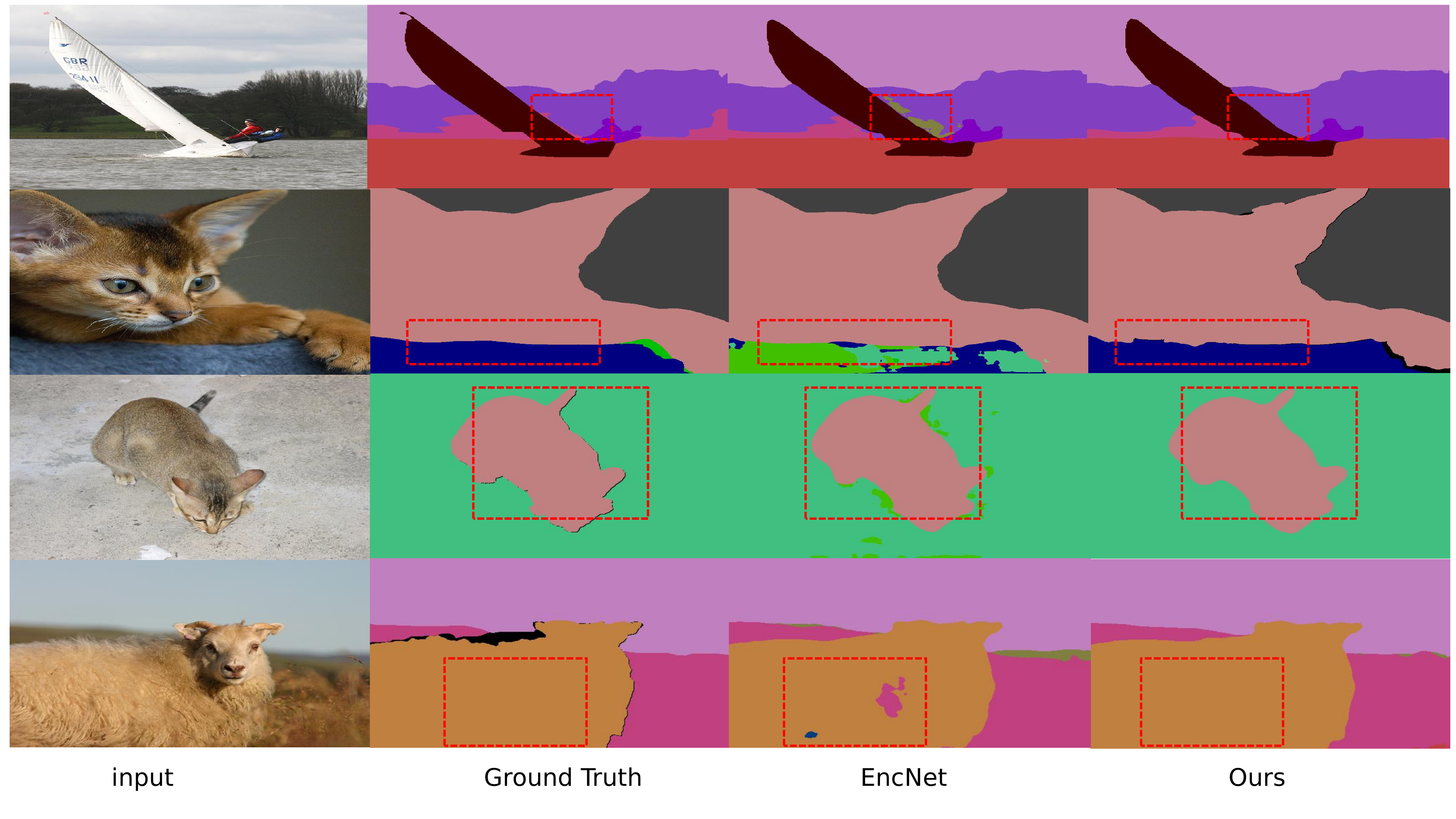}
	\caption{
		Comparison of our results on Pascal Context to the state-of-the-art EncNet~\cite{encodingnet}.
		Note that our results are more consistent and have fewer artifacts. Best view in color and zoom in.}
	\label{fig:pcontext_res}
\end{figure}

\vspace{1mm}
\noindent
\textbf{Results on Pascal Context:}
This dataset provides detailed semantic labels for the whole scenes\cite{pcontext-data}. It contains 4998 images for training and 5105 images for validation. We train the network for 100 epochs with a batch size of 16, a crop size of 480. For evaluation, we perform multi-scale testing with the flip operation, which boosts the results by about $1.2\%$ in mIoU. Fig.~\ref{fig:pcontext_res} shows the results of our method and EncNet. Compared with EncNet~\cite{encodingnet}, our method achieves better consistent results on the object inner parts, benefited from better reasoned features. Tab.~\ref{tab:seg_results_other}(a) reports results on Pascal Context. With ResNet-101 as the backbone,  our method achieves 54.7\% in mIoU with multi-scale inference, surpassing state-of-the-art alternatives by a large margin. Additionally, using the ResNet-50 backbone, we achieve 50.8\% mIoU, which also outperforms the previous work~\cite{DAnet} under the same setting.

\begin{figure}[t]
	\centering
	\vspace{-0.5\baselineskip}
	\includegraphics[width=1.0\linewidth]{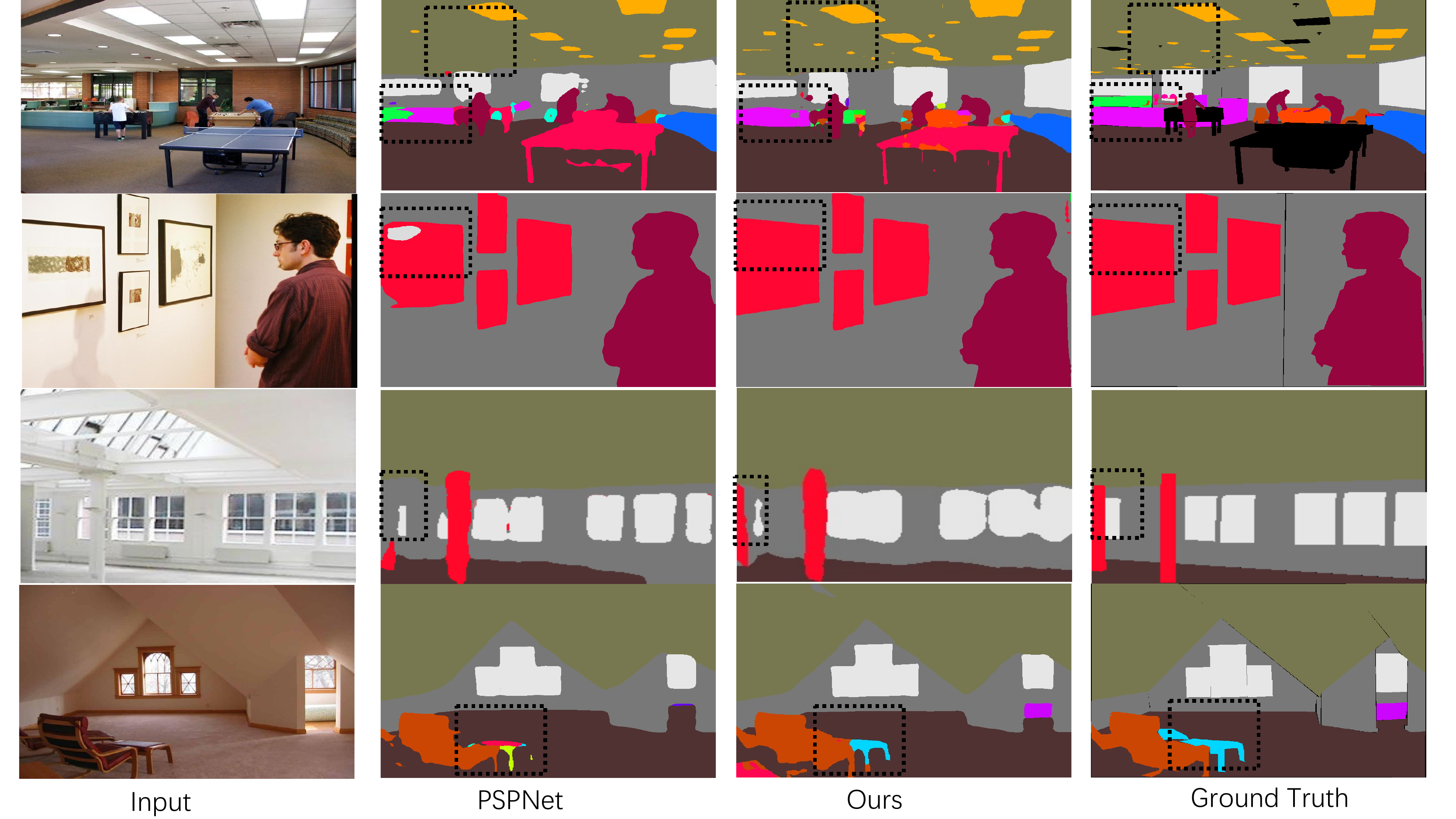}
	\caption{
		Comparison of our results with the state-of-the-art PSPNet \cite{pspnet} method on the ADE-20k dataset. The black boxes show that our method can get more consistent results and missing objects. Best view in color and zoom in.}
	\label{fig:ade_res}
\end{figure}

\vspace{1mm}
\noindent
\textbf{Results on ADE20K:} This is a more challenging scene parsing dataset annotated with 150 classes, which it contains 20k and 2k images for training and validation, respectively. We train the network for 120 epochs with a batch size of 16, a crop size of 512 and an initial learning rate of 1e-2. We perform multi-scale testing with the flip operation as the common setting in ~\cite{pspnet}. The visible results are shown in Fig.~\ref{fig:ade_res}. Compared to PSPNet~\cite{pspnet}, our method better handles inconsistent results and missing objects in the scene. In Tab.~\ref{tab:seg_results_other}(b), Both results using ResNet-50 and ResNet-101 backbone are reported. As shown in Tab.~\ref{tab:seg_results_other}(b), Our method with ResNet-101 achieves the best results and comparable results with ResNet-50 backbone. Our methods have less computation cost in the head part. Thus it results in an efficient inference.

\subsection{Results on MS COCO}
To verify our module's generality, we further conduct experiments on MS COCO~\cite{COCO_dataset} for more tasks, including object detection, instance segmentation and panoptic segmentation. The trainval set has about 115k images, the minival set has 5k images. We perform training on trainval set and report results on minival set. For the first two tasks, our model is based on the state-of-the-art method Mask R-CNN~\cite{maskrcnn} and its variants~\cite{cascade_rcnn,deformable}. For panoptic segmentation, we choose Panoptic FPN as our baseline~\cite{panoptic_fpn}.
We use open-source tools~\cite{mmdetection2018} to carry out all the experiments and report results on the MS COCO validation dataset. The GFlops are measured with 1200$\times$800 inputs.
Our models and all baselines are trained with the typical `1x' training schedule and setting from the public mmdetection~\cite{mmdetection2018} for all experiments on COCO.

\vspace{1mm}
\noindent
\textbf{Results on object detection and instance segmentation:} Tab.~\ref{tab:coco_results}(a) compares results of both object detection and instance segmentation with various backbone networks~\cite{resnext} and advanced method~\cite{cascade_rcnn}, where our method achieves consistently better performance on all backbones with much less computation cost compared with Non-Local blocks~\cite{Nonlocal}.

\vspace{1mm}
\noindent
\textbf{Results on panoptic segmentation:} Panoptic Segmentation~\cite{panoptic_seg} uses the PQ metric to capture the performance for all classes (stuff and things) in a unified way. We use the Panoptic-FPN~\cite{panoptic_fpn} as our baseline model and follow the standard settings in mmdetection. We re-implement the baseline model using mmdetection tools and achieve the similar results with original paper~\cite{panoptic_fpn}. The results are shown in Tab.~\ref{tab:coco_results}(b). Our method improves baseline and outperforms the non-local based methods through both overall evaluation and evaluations separated into thing and stuff, and the improvements are across both backbones, ResNet-50 and ResNet-101 with less computation cost. 

\begin{table}[!t]
	\caption{Experiments on COCO dataset. (a) Detection results on the COCO 2017 validation set. \textbf{R-50}: ResNet-50. \textbf{R-101}: ResNet-101. \textbf{X-101}: ResNeXt-101~\cite{resnext}. \textbf{NL}: Non-Local blocks~\cite{Nonlocal}.
	(b) Panoptic segmentation results on the COCO 2017 validation set.
	}
	\begin{minipage}{0.5\textwidth}
		\begin{minipage}{\dimexpr.95\linewidth}        
			\centering
			\resizebox{0.9\textwidth}{!}{%
				\begin{tabular}{c|c|c|c|c}
					\hline
					{Backbone} & {Detector} & {AP-box} & {AP-mask} &  {GFlops} \\
					\hline \hline
					R-50 & Mask-RCNN & 37.2  & 33.8 & 275.58 \\
					R-50 & +NL~\cite{Nonlocal} & 38.0    & 34.7  & +30.5 \\
					R-50 & \textbf{+SR}  & \textbf{38.4}    & \textbf{34.9} & +0.56 \\
					\hline
					R-101 & Mask-RCNN & 39.8 & 36.0 & 351.65 \\
					R-101 & +NL & 40.5  & 36.7 & +45.7 \\
					R-101 & \textbf{+SR} & \textbf{40.8}  & \textbf{36.9} & +1.32 \\
					\hline
					X-101 & Mask-RCNN & 41.2 & 37.3 & 355.37 \\
					X-101 & \textbf{+SR} & \textbf{42.0} & \textbf{37.8} & +1.32  \\
					\hline
					X-101 & Cascaded-Mask-RCNN & 44.7 & 38.3 & 519.90 \\
					X-101 & \textbf{+SR} & \textbf{45.5} & \textbf{39.0} & +1.32  \\
					\hline \hline
				\end{tabular}
			}\par
			\vspace{1mm}
			{\footnotesize(a) Experiments on COCO Object Detection and Instance Segmentation using various baseline models.}    
		\end{minipage}
		\vspace{2mm}
		
		\begin{minipage}{\dimexpr.95\linewidth}
			\centering
			\resizebox{0.9\textwidth}{!}{%
				\begin{tabular}{l|c|c|c|c|c}
					\hline
					Method & Backbone & PQ & PQ~(things) & PQ~(stuff) & {GFlops}  \\  
					\hline \hline
					Base & ResNet-50 & 39.0 & 45.9 & 28.7 & 270.8 \\
					+NL & ResNet-50 & 39.8 & 46.8   & 28.9 & +30.5 \\
					\textbf{+SR} & ResNet-50 & \textbf{40.3}  & \textbf{47.2} & \textbf{29.9} & +0.56 \\
					\hline
					Base & ResNet-101 & 40.3  & 47.5  & 29.5 & 346.87 \\
					+NL & ResNet-101 & 40.8  & 48.5  & 30.4 & +45.7 \\
					\textbf{+SR} & ResNet-101 & \textbf{41.8} & \textbf{48.7} & \textbf{31.3} & +1.32 \\
					\hline \hline
				\end{tabular}
			}\par
			\vspace{2mm}
			{\footnotesize(b) Experiments on COCO Panoptic Segmentation using PanopticFPN as baseline model. }    
		\end{minipage}
	\end{minipage}
	
	\label{tab:coco_results}
\end{table}

\begin{figure}[t]
	\centering
	\vspace{-0.5\baselineskip}
	\includegraphics[width=1.0\linewidth]{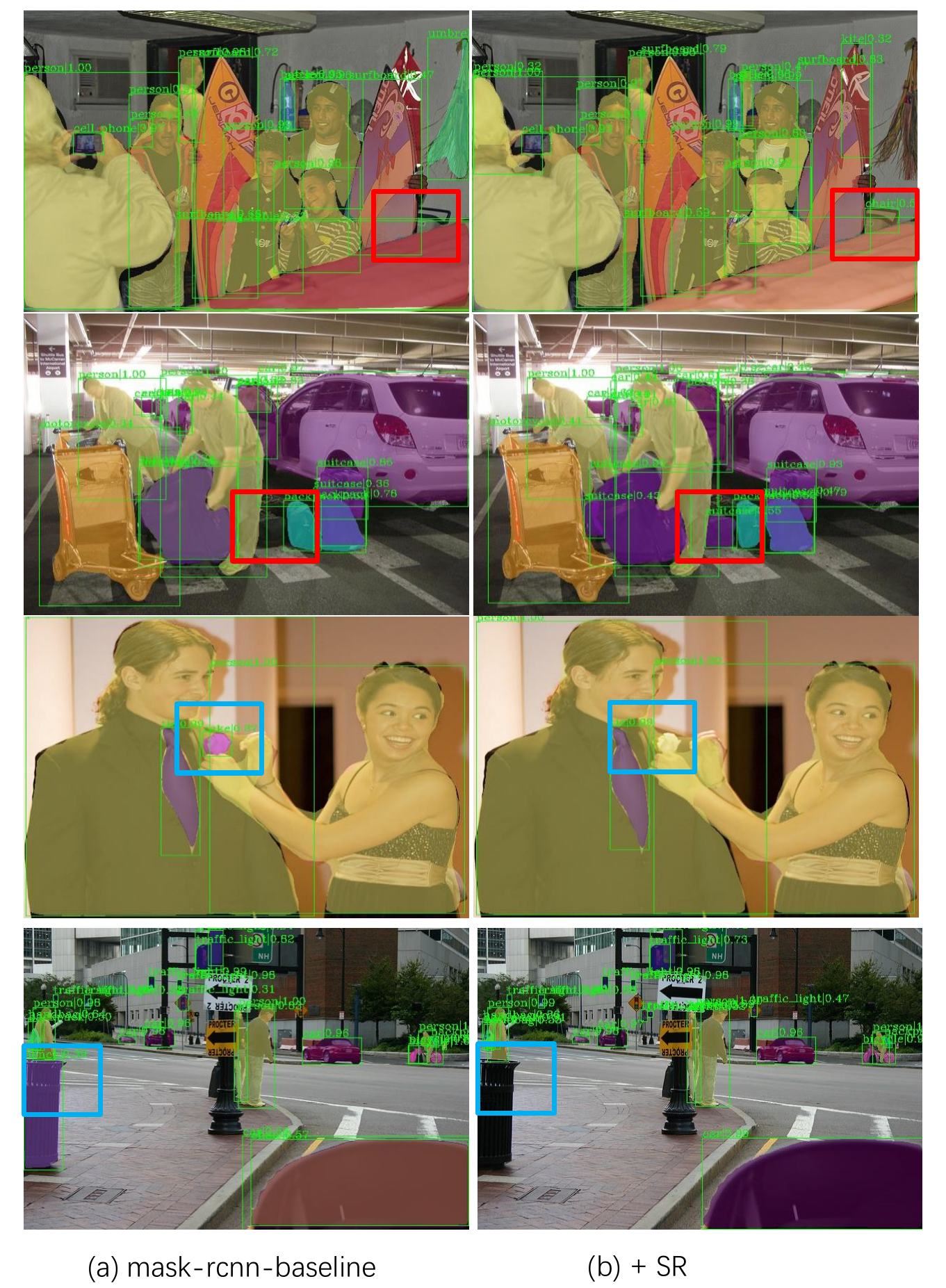}
	\caption{
		Comparison of our results on COCO with Mask-RCNN with ResNet101 backbone. Best view in color and zoom in.}
	\label{fig:coco_vis_res}
\end{figure}

\vspace{1mm}
\noindent
\textbf{Visualization Results on COCO:} Fig~\ref{fig:coco_vis_res} shows the qualitative results on COCO validation set. We use Mask-RCNN~\cite{Mask-RCNN} with ResNet101 as the baseline model.
The first two rows show our SR module can handle small missing objects (red boxes) while the last two rows show our method can also avoid false positives (blue boxes) in the scene.

\begin{table}[!t]
	\setlength\tabcolsep{3.6mm}
	\centering	
		\caption{Experiments Results on ImageNet. Our method achieves better Top-1 accuracy with lower parameter and computation cost on strong baseline settings. r50 means ResNet50 as backbone while r101 means ResNet101 as backbone.
		}
	\begin{tabular}{l|c|c|c}
		\hline
		Method & Top-1  & Params & GFLOPs\\ 
		\hline \hline
		ResNet-r50 & 77.5 & 25.56M & 4.122 \\
		SENet-r50~\cite{senet} & 77.8 & 28.09M & 4.130 \\
		CBAM-r50~\cite{cbam} & 78.0 & 28.09M & 4.139	\\
		\textbf{SRNet-r50} & \textbf{78.1} &  \underline{25.64M} & \underline{4.127} \\
		\hline
		ResNet-r101 & 78.7  &  44.55M & 7.849 \\
		SENet-r101~\cite{senet} & 79.1 & 49.33M & 7.863 \\
		\textbf{SRNet-r101}  & \textbf{79.3} & \underline{44.70M} & \underline{7.858} \\
		\hline
		\hline
	\end{tabular}

	\label{tab:imagenet_results}
	\vspace{0mm}
\end{table}

\subsection{Extension on ImageNet classification}

{We also perform experiments for image classification on ImageNet dataset~\cite{imagenet}, where SENet~\cite{senet} is used as our baseline model. To be noted, we only verify the effectiveness and generalization of SR framework for classification task.}
Our model is designed by replacing fully-connected layer with our proposed graph reasoning module, where the global hadamard pooling is not used for both saving computation and fair comparison with SEnet. All models are trained under the same setting, and results are shown in Table~\ref{tab:imagenet_results}. All networks are trained following the same strategy as~\cite{resnet} using Pytorch~\cite{pytorch} with 8 GTX 1080Ti GPUs. In particular, cosine learning rate schedule with warm up strategy is adopt~\cite{bag_of_trick}, and weight decay is set to 1e-4. SGD with mini-batch size 256 is used for updating weights. Top-1 and top-5 classification accuracy on validation set using single $224 \times 224$ central crop are reported for performance comparison. Due to the usage of cosine learning rate schedule~\cite{bag_of_trick}, our baseline models on ImageNet are higher than the original paper~\cite{resnet,senet}. Compared with both SENet and CBAM~\cite{cbam}, our SRNet improves the strong baseline SENet~\cite{senet} by 0.3 in Top-1 accuracy with much less parameter and GFlops. Our method leads to higher accuracy with fewer parameters and FLOPs, which demonstrates both effectiveness and efficiency of the proposed method.

\section{Conclusion}
This paper proposes a novel Squeezing and Reasoning framework for highly efficient deep feature representation learning for the scene understanding tasks. It learns to squeeze the feature to a node graph space where each node represents an abstract semantic concept while both memory and computation costs are significantly reduced.  Extensive experiments demonstrate that our method can establish considerable results on semantic segmentation while keeping efficiency compared with previous the-state-of-the-art models. It also shows consistent improvement with respect to strong baselines over instance segmentation and panoptic segmentation with much less computation. It also verified to be effective on image classification task and better results over SENet. The further work can be exploring cross layer reasoning over entire network.


%


\section*{Acknowledgment}
We gratefully acknowledge the support of SenseTime Research for providing the computing resources. Z. Lin is supported by NSF China (grant no.s 61625301 and 61731018) and Major Scientific Research Project of Zhejiang Lab (grant no.s 2019KB0AC01 and 2019KB0AB02). This work was also supported by the National Key Research and Development Program of China (No.2020YFB2103402).

\ifCLASSOPTIONcaptionsoff
  \newpage
\fi



{
\bibliographystyle{IEEEtran}
\bibliography{IEEEabrv,egbib}
}
\end{document}